\documentclass[runningheads]{llncs}

\usepackage{eccv}

\usepackage{eccvabbrv}

\usepackage{graphicx}
\usepackage{booktabs}
\usepackage{graphicx}    % \resizebox
\usepackage{xcolor}      % \textcolor[HTML]{...}
\usepackage[accsupp]{axessibility}  % Improves PDF readability for those with disabilities.

\usepackage{wrapfig}
\usepackage{booktabs}
\usepackage{float}
\usepackage{graphicx}
\usepackage{makecell}
\usepackage{adjustbox}
\usepackage{tikz}
\usepackage{multirow}

\usepackage{subcaption}
\usepackage{amsmath}
\usepackage{amssymb}
\usepackage{booktabs}
\usepackage{adjustbox}
\usepackage{caption}

\usepackage{multirow}
\usepackage{multicol}
\usepackage{upgreek}
\usepackage{array}

\usepackage{tabularx}
\usepackage{algorithm}
\usepackage{algorithmic}

\usepackage{amsmath,amsfonts}

\usepackage{amssymb}
\usepackage{amsmath}
\usepackage{mathtools}
\usepackage{amsfonts}
\usepackage{bbding}
\usepackage{float}
\newcommand{\mycomment}[1]{\hfill {// #1}}
\newcommand{\ourall}{\text{DEFAR}}
\newcommand{\ourdri}{\text{ADR}}
\newcommand{\ourfre}{\text{FC}}

\newsavebox{\lefttablebox}

\usepackage{float}
\usepackage{caption}


\usepackage{hyperref}
\hypersetup{hidelinks}

\usepackage{orcidlink}

\usepackage{adjustbox} \usepackage{calc}
\begin{document}

\title{Exposure Bias Can Alleviate Itself via Directional and Frequency Rectification in Flow Matching} 
% Exposure Bias can Alleviate Itself via Direction and Frequency Rectification in Flow Matching
% Exposure Bias can Alleviate Itself, via direction and frequency rectification in Flow Matching
% Exposure Bias can Dynamically Alleviate Itself: Anti-Drift Rectification and Frequency Compensation for Exposure Bias Mitigation in Flow Matching
% Self-Alleviating Exposure Bias in Flow Matching via Anti-Drift and Frequency Compensation

\titlerunning{DEFAR for Exposure Bias Alleviation.}

\author{Guanbo Huang\inst{1,2}$^{*,\ddagger}$ \and
Jingjia Mao\inst{1,2}$^{*,\ddagger}$ \and
Fanding Huang\inst{1}$^{*}$ \and
Fengkai Liu\inst{1} \and
Xiangyang Luo\inst{1} \and
Yaoyuan Liang\inst{1} \and
Jiasheng Lu\inst{2} \and
Xiaoe Wang\inst{2} \and
Pei Liu\inst{2} \and
Ruiliu Fu\inst{2}$^{\dagger}$ \and
Ruqi Huang\inst{1}$^{\dagger}$ \and
Shao-Lun Huang\inst{1}$^{\dagger}$
}

\authorrunning{G.~Huang, J.~Mao, F.~Huang et al.}

\institute{Tsinghua Shenzhen International Graduate School, Tsinghua University\and
Central Media Technology Institute, Huawei\\}

\maketitle

\begingroup
\renewcommand{\thefootnote}{}
\footnotetext{$^{*}$ Equal contributions.\\
$^{\dagger}$ Corresponding authors: Ruiliu Fu, Ruqi Huang and Shao-Lun Huang.\\
$^{\ddagger}$ This work was conducted during the internship at Central Media Technology Institute, Huawei.}
\endgroup

\newcommand{\sectionwithouttoc}[1]{%
  \refstepcounter{section}%
  \sectionmark{#1}%
  \section*{\thesection\quad #1}%
}

\newcommand{\subsectionwithouttoc}[1]{%
  \refstepcounter{subsection}%
  \subsectionmark{#1}%
  \subsection*{\thesubsection\quad #1}%
}

% \begin{abstract}
%   The abstract should concisely summarize the contents of the paper. 
%   While there is no fixed length restriction for the abstract, it is recommended to limit your abstract to approximately 150 words.
%   Please include keywords as in the example below. 
%   This is required for papers in LNCS proceedings.
%   \keywords{First keyword \and Second keyword \and Third keyword}
% \end{abstract}

\def\TabA{
\begin{table*}[t]
\centering
\caption{\textbf{Quantitative comparison on conditional datasets.} Class-conditional (without/with CFG following SiT~\cite{SiT2024}) generation on ImageNet-256 and CIFAR-10. All entries are reported using 50 sampling steps (NFE). \textcolor[HTML]{CC0000}{Red} highlights improvements by our method, and \textcolor[HTML]{009900}{green} indicates degradations compared to \textcolor[HTML]{A0A0A0}{gray} baseline. Under a comparable cost (e.g., 144h vs. 166h on ImageNet), \ourall{} outperforms the baselines.}
\renewcommand{\arraystretch}{1.1}
\resizebox{\textwidth}{0.08\textheight}{
\begin{tabular}{l c c c c ccccc c c c c c c c c c c c c cccc}
\toprule
\multirow{2.5}{*}{\textbf{Method}} & \multirow{2.5}{*}{\textbf{Backbone}} &\multirow{2.5}{*}{\textbf{Params}} & \multicolumn{12}{c}{\textbf{ImageNet-256 (conditional)}} & \multicolumn{11}{c}{\textbf{CIFAR-10 (conditional)}}\\  
\cmidrule(lr){4-14} \cmidrule(lr){15-26} 
 &&&Iters & Hour & FID$\downarrow$&& sFID$\downarrow$ && IS$\uparrow$ && Pre.$\uparrow$ && Rec.$\uparrow$ && Iters & Hour & FID$\downarrow$&& sFID$\downarrow$ && IS$\uparrow$ && Pre.$\uparrow$ && Rec.$\uparrow$ \\ 
\midrule
SiT (origin) \cite{SiT2024} & SiT-B/4  & 131M & 500k & 166 & 61.64/31.64 && 12.20/8.44 && 24.67/55.65 && 0.387/0.542 && 0.581/0.524 && 250k & 11 & 14.09/9.36  && 6.28/5.73 && 8.22/8.90 && 0.617/0.676 && 0.538/0.500\\
 &&&&&  \textcolor[HTML]{A0A0A0}{+0.0}/\textcolor[HTML]{A0A0A0}{+0.0} && \textcolor[HTML]{A0A0A0}{+0.0}/\textcolor[HTML]{A0A0A0}{+0.0}&& \textcolor[HTML]{A0A0A0}{+0.0}/\textcolor[HTML]{A0A0A0}{+0.0}&& \textcolor[HTML]{A0A0A0}{+0.0}/\textcolor[HTML]{A0A0A0}{+0.0} && \textcolor[HTML]{A0A0A0}{+0.0}/\textcolor[HTML]{A0A0A0}{+0.0}&&&& \textcolor[HTML]{A0A0A0}{+0.0}/\textcolor[HTML]{A0A0A0}{+0.0}&&\textcolor[HTML]{A0A0A0}{+0.0}/\textcolor[HTML]{A0A0A0}{+0.0}&& \textcolor[HTML]{A0A0A0}{+0.0}/\textcolor[HTML]{A0A0A0}{+0.0}&&\textcolor[HTML]{A0A0A0}{+0.0}/\textcolor[HTML]{A0A0A0}{+0.0}&&\textcolor[HTML]{A0A0A0}{+0.0}/\textcolor[HTML]{A0A0A0}{+0.0}\\
IP \cite{IP2023} & SiT-B/4 & 131M & 500k & 166 & 60.45/29.84 && 12.49/8.80 && 26.05/60.20 && \textbf{0.404}/\textbf{0.563} && 0.578/0.523 && 250k& 11& 13.63/8.76 && 6.43/5.58 && 8.18/8.97  && \underline{0.624}/\underline{0.685} && 0.539/0.500\\
SDSS \cite{multistep2024} & SiT-B/4  & 131M &500k& 228 &61.25/30.29 && 13.23/8.95 && 25.81/58.69 && 0.387/0.549 && 0.588/\underline{0.534} && 250k &15 & 13.27/8.27 && 5.97/5.41 && 8.21/8.91 && 0.612/0.679 && 0.550/0.506\\
MDSS (4steps) \cite{multistep2024} & SiT-B/4 & 131M &500k& 228 &63.66/31.90 && 14.38/10.05 && 24.72/56.89 && 0.382/0.544 && 0.583/0.527 && 250k & 15 &12.73/8.45 && 6.29/5.93 && 8.25/\underline{9.06} && \textbf{0.631}/\textbf{0.689} && 0.529/0.484\\
\midrule
\ourall{} w/o \ourfre{} & SiT-B/4  & 131M & 500k & 228 &59.77/29.69 && 12.77/9.00 && 25.81/\underline{59.25} && 0.377/0.539 && \textbf{0.598}/\textbf{0.549} && 200k & 12 & 11.61/7.62 && 6.07/6.21 && \underline{8.30}/9.05 && 0.599/0.669 && \underline{0.556}/\underline{0.507}\\
\ourall{} w/o \ourdri{} & SiT-B/4  & 131M & 500k &289 &\underline{57.17}/\underline{28.71} && 10.65/8.25 && \underline{26.55}/\textbf{60.43} && \underline{0.398}/\underline{0.557} && 0.590/0.532 && 200k & 16 & \underline{11.25}/\textbf{7.53} && 5.71/5.53 && \underline{8.30}/\textbf{9.10} && 0.601/0.677 && \textbf{0.557}/0.506\\
\ourall{} & SiT-B/4 & 131M &250k& 144 & 60.40/29.81 && \textbf{9.05}/\textbf{7.35} && 26.21/58.81 && 0.375/0.531 && 0.586/0.532 && 120k & 10 & 12.01/7.73 && \underline{5.46}/\textbf{5.13} && 8.27/9.01 && 0.605/0.667 && 0.555/\textbf{0.514}\\
&&&&& \textcolor[HTML]{CC0000}{-1.24}/ \textcolor[HTML]{CC0000}{-1.83} && \textcolor[HTML]{CC0000}{-3.15}/ \textcolor[HTML]{CC0000}{-1.09} && \textcolor[HTML]{CC0000}{+1.54}/\textcolor[HTML]{CC0000}{+3.16} && \textcolor[HTML]{009900}{-0.012}/\textcolor[HTML]{009900}{-0.011} && \textcolor[HTML]{CC0000}{+0.005}/\textcolor[HTML]{CC0000}{+0.008} &&&& \textcolor[HTML]{CC0000}{-2.08}/\textcolor[HTML]{CC0000}{-1.63} && \textcolor[HTML]{CC0000}{-0.82}/\textcolor[HTML]{CC0000}{-0.60}&& \textcolor[HTML]{CC0000}{+0.05}/\textcolor[HTML]{CC0000}{+0.11}&& \textcolor[HTML]{009900}{-0.012}/\textcolor[HTML]{009900}{-0.009}&& \textcolor[HTML]{CC0000}{+0.017}/\textcolor[HTML]{CC0000}{+0.014}\\
\ourall{} & SiT-B/4 & 131M &500k& 289 & \textbf{56.39}/\textbf{28.34} && \underline{9.50}/\underline{7.84} && \textbf{26.62}/58.94  && 0.386/0.551 && \underline{0.594}/0.524 && 180k &15  & \textbf{11.22}/\underline{7.56} && \textbf{5.32}/\underline{5.21} && \textbf{8.31}/\underline{9.06} && 0.618/0.682 && 0.553/0.500\\
\bottomrule
\end{tabular}
}
\label{tab:major_conditional}
\end{table*}
}

\def\TabAtwo{%
\begin{table*}[t]
\centering
\caption{\textbf{Unconditional generation and complementary mitigation strategies.}
\textbf{Left:} Quantitative results on unconditional CIFAR-10 and CelebA-64 (50 NFE), where \ourall{} (120k) surpasses baselines under a comparable training budget.
\textbf{Right:} Complementarity with two exposure-bias mitigation paradigms, DG on conditional ImageNet-256 and mini-batch OT coupling on unconditional CIFAR-10.} % Modified: reframe rebuttal additions as complementary exposure-bias baselines.
\label{tab:major_unconditional}
\newsavebox{\TabAtwoLeftBox}
\newsavebox{\TabAtwoRightBox}
\def\TabAtwoPanelHeight{0.06\textheight}
\savebox{\TabAtwoLeftBox}{%
{\renewcommand{\arraystretch}{0.8}%
\begin{tabular}{l c c c c c c c c c c c c c c c}
\toprule
\multirow{2.5}{*}{\textbf{Method}} & \multirow{2.5}{*}{\textbf{Backbone}} &\multirow{2.5}{*}{\textbf{Params}}&\multirow{2.5}{*}{\textbf{Iters}} & \multicolumn{6}{c}{\textbf{CIFAR-10}} & \multicolumn{6}{c}{\textbf{CelebA-64}}\\ 
\cmidrule(lr){5-10} \cmidrule(lr){11-16} 
 &&&&Hour& FID$\downarrow$& sFID$\downarrow$ & IS$\uparrow$ & Pre.$\uparrow$ & Rec.$\uparrow$ &Hour& FID$\downarrow$& sFID$\downarrow$ & IS$\uparrow$ & Pre.$\uparrow$ & Rec.$\uparrow$ \\ 
\midrule
SiT (origin) \cite{SiT2024} & SiT-B/4 & 131M &250k & 12 & 17.41 & 6.19 & 7.82 & 0.586 & 0.539 &36& 7.04& 6.88 & 2.76 & 0.626 & 0.521\\
 &&&&&  \textcolor[HTML]{A0A0A0}{+0.0} & \textcolor[HTML]{A0A0A0}{+0.0}& \textcolor[HTML]{A0A0A0}{+0.0}& \textcolor[HTML]{A0A0A0}{+0.0}& \textcolor[HTML]{A0A0A0}{+0.0} && \textcolor[HTML]{A0A0A0}{+0.0}&\textcolor[HTML]{A0A0A0}{+0.0}& \textcolor[HTML]{A0A0A0}{+0.0} & \textcolor[HTML]{A0A0A0}{+0.0}&\textcolor[HTML]{A0A0A0}{+0.0}\\
IP \cite{IP2023} & SiT-B/4 & 131M &250k & 12 & 15.95 & 6.65 & \underline{7.97} & 0.583 & 0.556 &36 & 5.93 & 6.57 & 2.72 & 0.640 & \underline{0.543}\\
SDSS \cite{multistep2024} & SiT-B/4 & 131M &250k & 17 & 16.02 & 6.37 & 7.70 & \underline{0.590} & 0.542 &60& 6.12 & 8.01 & 2.71 & \underline{0.671} & 0.476\\
MDSS(4steps) \cite{multistep2024} & SiT-B/4 & 131M &250k &17 & 16.10 & 6.59 & 7.91 & \textbf{0.594} & 0.547 & 60 & 6.10 & 8.29 & 2.91 & 0.623 & 0.527 \\
\midrule
\ourall{} w/o \ourfre{} & SiT-B/4 & 131M & 250k & 17 & \underline{15.92} & 6.45 & 7.83 & 0.585 & 0.554 &60 & 6.83 & 6.47 & 2.67 & \textbf{0.676} & 0.498\\
\ourall{} w/o \ourdri{} & SiT-B/4 & 131M &250k & 23 & 15.99 & \underline{5.99} & \textbf{8.07} & 0.580 & 0.562 & 80 & \underline{4.96} & \underline{6.26} & \textbf{2.94} & 0.641 & 0.541\\
\ourall{} & SiT-B/4 & 131M &120k &11 & 15.93 & \underline{5.99} & 7.86 & 0.567 & \textbf{0.567} & 36 & 5.91 & \textbf{6.13} & 2.77 & 0.634 & 0.521\\
 &&&&& \textcolor[HTML]{CC0000}{-1.48} & \textcolor[HTML]{CC0000}{-0.20} & \textcolor[HTML]{CC0000}{+0.04} & \textcolor[HTML]{009900}{-0.019} & \textcolor[HTML]{CC0000}{+0.028} &&  \textcolor[HTML]{CC0000}{-1.13} &\textcolor[HTML]{CC0000}{-0.75}& \textcolor[HTML]{CC0000}{+0.01} & \textcolor[HTML]{CC0000}{+0.008}&\textcolor[HTML]{CC0000}{+0.000}\\
\ourall{} & SiT-B/4 & 131M &250k & 23 & \textbf{14.82} & \textbf{5.88} & 7.79 & 0.570 & \underline{0.563} & 80 & \textbf{4.53} & 6.41 & \underline{2.92} & 0.629 & \textbf{0.551}\\
\bottomrule
\end{tabular}%
}}
\savebox{\TabAtwoRightBox}{%
{\renewcommand{\arraystretch}{0.78}%
\begin{minipage}{\textwidth}
\resizebox{\linewidth}{!}{%
\begin{tabular}{lcccccccc}
\toprule
\multicolumn{9}{l}{\textit{Discriminator guidance~\cite{kim2023refining} (conditional ImageNet-256 with SiT-B/4):}}\\[1.2ex] % Modified: more paper-logical block name.
\textbf{Method} & \textbf{Cls./Disc.} & FID$\downarrow$ & sFID$\downarrow$ & IS$\uparrow$ & Pre.$\uparrow$ & Rec.$\uparrow$ & $\Updelta$FID & $\Updelta$IS\\
\midrule
SiT & - & 61.64 & 12.20 & 24.67 & \underline{0.387} & 0.581 & \textcolor[HTML]{A0A0A0}{+0.00} & \textcolor[HTML]{A0A0A0}{+0.00}\\
% SiT + \ourdri{} & - & 60.76 & 12.44 & 25.33 & 0.372 & \textbf{0.593} & \textcolor[HTML]{CC0000}{-0.88} & \textcolor[HTML]{CC0000}{+0.66}\\
\ourall{} & - & \underline{60.40} & \underline{9.05} & \underline{26.21} & 0.375 & 0.586 & \textcolor[HTML]{CC0000}{-1.24} & \textcolor[HTML]{CC0000}{+1.54}\\
SiT-G++ & ADM/U-Net$_{\mathrm{shallow}}$ & 60.92 & 11.10 & 25.08 & \textbf{0.388} & 0.579 & \textcolor[HTML]{CC0000}{-0.72} & \textcolor[HTML]{CC0000}{+0.41}\\
\ourall{}-G++ & ADM/U-Net$_{\mathrm{shallow}}$ & \textbf{59.03} & \textbf{8.84} & \textbf{26.49} & 0.374 & \underline{0.590} & \textcolor[HTML]{CC0000}{-2.61} & \textcolor[HTML]{CC0000}{+1.82}\\
\end{tabular}%
}\par
\resizebox{\linewidth}{!}{%
\begin{tabular}{lccccccc}
\toprule
\multicolumn{8}{l}{\textit{Mini-batch OT coupling~\cite{tong2024improving} (unconditional CIFAR-10):}}\\[1.2ex] % Modified: more paper-logical block name.
\textbf{Method} & \textbf{Backbone} & \textbf{Hour} & \textbf{Iters} & FID$_{100}\downarrow$ & FID$_{1000}\downarrow$ & Avg. FID$\downarrow$ & $\Updelta$Avg. FID\\
\midrule
OT-FM & U-Net & 20 & 400k & 4.643{\scriptsize$\pm$0.035} & 3.824{\scriptsize$\pm$0.037} & 4.233{\scriptsize$\pm$0.027} & \textcolor[HTML]{A0A0A0}{+0.000{\scriptsize$\pm$0.000}}\\
OT-FM + \ourall{} & U-Net & 20 & 200k & \underline{4.397{\scriptsize$\pm$0.029}} & \underline{3.716{\scriptsize$\pm$0.017}} & \underline{4.056{\scriptsize$\pm$0.018}} & \textcolor[HTML]{CC0000}{-0.177{\scriptsize$\pm$0.023}}\\
OT-CFM & U-Net & 20 & 400k & 4.446{\scriptsize$\pm$0.037} & 3.744{\scriptsize$\pm$0.029} & 4.095{\scriptsize$\pm$0.022} & \textcolor[HTML]{CC0000}{-0.138{\scriptsize$\pm$0.033}}\\
OT-CFM + \ourall{} & U-Net & 20 & 200k & \textbf{4.260{\scriptsize$\pm$0.047}} & \textbf{3.660{\scriptsize$\pm$0.037}} & \textbf{3.960{\scriptsize$\pm$0.034}} & \textcolor[HTML]{CC0000}{-0.273{\scriptsize$\pm$0.029}}\\
\bottomrule
\end{tabular}%
}
\end{minipage}%
}}
\resizebox{0.57\textwidth}{\TabAtwoPanelHeight}{\usebox{\TabAtwoLeftBox}}\hfill
\resizebox{0.41\textwidth}{\TabAtwoPanelHeight}{\usebox{\TabAtwoRightBox}}
\end{table*}
}

\def\TabBTwo{%
\begin{table*}[t]
\centering
\caption{\textbf{Results on ImageNet-256 conditional generation with CFG.}
Scalability analysis of \ourall{} versus SiT across different model sizes and training iterations. All entries are reported using 50 sampling steps (NFE).}
\label{tab:scalability}
\renewcommand{\arraystretch}{0.8}
\resizebox{\textwidth}{!}{%
    \begin{tabular}{l c c c c c c c c c}
    \toprule
    \multirow{2.5}{*}{\textbf{Method}} & \multirow{2.5}{*}{\textbf{Backbone}} &\multirow{2.5}{*}{\textbf{Params}} &\multirow{2.5}{*}{\textbf{Iters}} &\multirow{2.5}{*}{\textbf{Hour}} & \multicolumn{5}{c}{\textbf{ImageNet-256}}\\ 
    \cmidrule(lr){6-10}
     &&&&& FID$\downarrow$ & sFID$\downarrow$ & IS$\uparrow$& Pre.$\uparrow$& Rec.$\uparrow$\\ 
    \midrule
    SiT \cite{SiT2024} & SiT-B/4 & 131M& 500k & 166 & 61.64/31.64 & 12.20/8.44 & 24.67/55.65  & 0.387/0.542 & 0.581/0.524\\
    \ourall{} & SiT-B/4 & 131M &250k& 144 & 60.40/29.81 & 9.05/7.35 & 26.21/58.81 & 0.375/0.531 & 0.586/0.532\\
    &&&&&\textcolor[HTML]{CC0000}{-1.24}/\textcolor[HTML]{CC0000}{-1.83} &\textcolor[HTML]{CC0000}{-3.15}/\textcolor[HTML]{CC0000}{-1.09}&\textcolor[HTML]{CC0000}{+1.54}/\textcolor[HTML]{CC0000}{+3.16}&\textcolor[HTML]{009900}{-0.012}/\textcolor[HTML]{009900}{-0.011}&\textcolor[HTML]{CC0000}{+0.005}/\textcolor[HTML]{CC0000}{+0.008}\\
    \ourall{} & SiT-B/4 & 131M &500k & 289 & 56.39/28.34  & 9.50/7.84 & \textbf{26.62}/\textbf{58.94} & 0.386/0.551 & 0.594/0.524\\
    \midrule
    SiT \cite{SiT2024} & SiT-M/2 & 308M & 700k & 311 & 30.57/8.23 & 10.90/6.13 & 58.66/154.71 & 0.547/0.750 & 0.654/0.531\\
    \ourall{} & SiT-M/2 & 308M & 350k & 330 & 29.46/7.85 & 5.60/4.64 & 59.01/156.27 & 0.549/0.751 & 0.639/0.518\\
    &&&&&\textcolor[HTML]{CC0000}{-1.11}/\textcolor[HTML]{CC0000}{-0.38} &\textcolor[HTML]{CC0000}{-5.30}/\textcolor[HTML]{CC0000}{-1.49}&\textcolor[HTML]{CC0000}{+0.35}/\textcolor[HTML]{CC0000}{+1.56}&\textcolor[HTML]{CC0000}{+0.002}/\textcolor[HTML]{CC0000}{+0.001}&\textcolor[HTML]{009900}{-0.015}/\textcolor[HTML]{009900}{-0.013}\\
    \ourall{} & SiT-M/2 & 308M & 700k & 661 & 27.58/7.30 & 7.35/4.96 & \textbf{62.09}/\textbf{159.76} & 0.546/0.752 & 0.658/0.531\\
    \midrule
    SiT \cite{SiT2024} & SiT-XL/2 & 675M & 700k & 2260 & 25.85/5.88 & 7.63/4.73 & 68.93/183.58 & 0.570/0.762 & 0.659/0.541\\
    \ourall{} & SiT-XL/2 & 675M & 350k & 2300 & 24.93/5.74 & 5.71/4.71 & 71.63/188.75 & 0.580/0.775 & 0.651/0.532\\
    &&&&&\textcolor[HTML]{CC0000}{-0.92}/\textcolor[HTML]{CC0000}{-0.14} &\textcolor[HTML]{CC0000}{-1.92}/\textcolor[HTML]{CC0000}{-0.02}&\textcolor[HTML]{CC0000}{+2.70}/\textcolor[HTML]{CC0000}{+5.17}&\textcolor[HTML]{CC0000}{+0.010}/\textcolor[HTML]{CC0000}{+0.013}&\textcolor[HTML]{009900}{-0.008}/\textcolor[HTML]{009900}{-0.009}\\
    \ourall{} & SiT-XL/2 & 675M &700k & 4640 & 18.45/4.14 & 6.34/4.95 & \textbf{77.80}/\textbf{199.59} & 0.621/0.810 & 0.644/0.529\\
    \bottomrule
    \end{tabular}%
}
\end{table*}
}

\newsavebox{\rightboxC}
\newsavebox{\leftboxC}

\def\TableHeightC{1.2cm}   

\def\LeftWidthC{0.45\textwidth}  
\def\RightWidthC{0.53\textwidth} 
% ==============================================

\def\TabC{%
\begin{table*}[t]
\centering
\caption{\textbf{Architectural compatibility and sampling robustness.} 
\textbf{Left:} Integration into REPA and DDT. 250-NFE generation under CFG, with and without \ourall{}. Models are fine-tuned on ImageNet-512 from official ImageNet-256 checkpoints. 
\textbf{Right:} Generation across varying NFEs without CFG on ImageNet-256.}
\label{tab:compatibility_and_nfe}
\savebox{\leftboxC}{%
\begin{tabular}{l c c c c c c c c}
    \toprule
    \multirow{2.5}{*}{\textbf{Model}} &\multirow{2.5}{*}{\textbf{Params}} &\multirow{2.5}{*}{\textbf{Iters}}  &\multirow{2.5}{*}{\textbf{Hour}} & \multicolumn{5}{c}{\textbf{ImageNet-512}}\\ 
    \cmidrule(lr){5-9}
     &&&& FID$\downarrow$& sFID$\downarrow$ & IS$\uparrow$ & Pre.$\uparrow$ & Rec.$\uparrow$\\ 
    \midrule
    % REPA-XL/2  & 675M & && 2.09 &4.19 &  & 0.830 & 0.580\\
    % REPA-XL/2 + \ourall{}  & 675M & 50k & & 1.96 & 4.21 &  & 0.841 & 0.573\\
    REPA-XL/2~\cite{repa2025}  & 675M & 100k & 400 &3.99& 11.76 & 248.60 & 0.796 & \textbf{0.595} \\
    \  + \textbf{\ourall{}}  & 675M & 45k & 410 & \textbf{3.46} & \textbf{9.12} & \textbf{255.26} & \textbf{0.798} & 0.582\\
    &&&&\textcolor[HTML]{CC0000}{-0.53} &\textcolor[HTML]{CC0000}{-2.64}&\textcolor[HTML]{CC0000}{+6.66}&\textcolor[HTML]{CC0000}{+0.002}&\textcolor[HTML]{009900}{-0.013}\\
     % &&&&  \textcolor[HTML]{A0A0A0}{+0.0\%} & \textcolor[HTML]{A0A0A0}{+0.0\%} &\textcolor[HTML]{A0A0A0}{+0.0\%} & \textcolor[HTML]{A0A0A0}{+0.0\%} &\textcolor[HTML]{A0A0A0}{+0.0\%} \\
    % FlowDCN-XL/2  & 618M & 2.09 &5.69 & 56.66 & 0.509 & 0.669\\
    % FlowDCN-XL/2 + \ourall{} & 618M & 2.09 &5.69 & 56.66 & 0.509 & 0.669\\
    % &&  \textcolor[HTML]{A0A0A0}{+0.0\%} & \textcolor[HTML]{A0A0A0}{+0.0\%} &\textcolor[HTML]{A0A0A0}{+0.0\%} & \textcolor[HTML]{A0A0A0}{+0.0\%} &\textcolor[HTML]{A0A0A0}{+0.0\%} \\
    \midrule
    DDT-XL/2~\cite{ddt2025}  & 675M & 100k &570 & 1.90 & 4.33 & 281.25 & 0.793 & \textbf{0.601} \\
    \ + \textbf{\ourall{}}  & 675M & 45k &580 & \textbf{1.82} & \textbf{4.21} & \textbf{285.42} & \textbf{0.799} & 0.596 \\
    &&&&\textcolor[HTML]{CC0000}{-0.08} &\textcolor[HTML]{CC0000}{-0.12}&\textcolor[HTML]{CC0000}{+4.17}&\textcolor[HTML]{CC0000}{+0.006}&\textcolor[HTML]{009900}{-0.005}\\
    % &&&&  \textcolor[HTML]{A0A0A0}{+0.0\%} & \textcolor[HTML]{A0A0A0}{+0.0\%} &\textcolor[HTML]{A0A0A0}{+0.0\%} & \textcolor[HTML]{A0A0A0}{+0.0\%} &\textcolor[HTML]{A0A0A0}{+0.0\%} \\
    \bottomrule
    \end{tabular}%
}

\savebox{\rightboxC}{%
\begin{tabular}{lcccccccccccc}
    \toprule
    \multirow{2.5}{*}{\textbf{Steps}} & \multicolumn{2}{c}{\textbf{SiT}} &\multicolumn{2}{c}{\textbf{IP}} & \multicolumn{2}{c}{\textbf{SDSS}} & \multicolumn{2}{c}{\textbf{MDSS}} & \multicolumn{2}{c}{\textbf{\ourall{}(250k)}}& \multicolumn{2}{c}{\textbf{\ourall{}(500k)}}\\ 
    \cmidrule(lr){2-3} \cmidrule(lr){4-5} \cmidrule(lr){6-7} \cmidrule(lr){8-9} \cmidrule(lr){10-11} \cmidrule(lr){12-13}
     & FID$\downarrow$ & IS$\uparrow$ & FID$\downarrow$ & IS$\uparrow$ & FID$\downarrow$ & IS$\uparrow$ & FID$\downarrow$ & IS$\uparrow$ & FID$\downarrow$ & IS$\uparrow$& FID$\downarrow$ & IS$\uparrow$\\ 
    \midrule
    30 & 63.63 & 24.70 & 61.25 & \textbf{26.21} & 63.41 & 25.80  & 65.93 & 24.78 & \underline{61.12} & 25.62 & \textbf{57.20} & \underline{25.84} \\
    \midrule
    50 & 61.64 & 24.67 & 60.45 & 26.05 & 61.25 & 25.81 & 63.66 & 24.72 &\underline{60.40} & \underline{26.21} & \textbf{56.39} & \textbf{26.62} \\
    \midrule
    100 & 60.42 & 24.48 & 58.26 & 26.01 & 59.92 & 25.78 & 61.33 &  24.64 & \underline{58.11} & \underline{26.09} & \textbf{55.89} & \textbf{26.51} \\
    \midrule
    250 & 59.84 & 24.30 & 57.75 & 25.83 & 59.23 & 25.64 & 61.33 & 24.48 & \underline{57.56} & \underline{25.91} & \textbf{55.66} & \textbf{26.33} \\
    \midrule
    500 & 59.70 & 24.24 & 57.58 & 25.76 & 59.06 & 25.56 & 61.10 & 24.41 & \underline{57.32} & \underline{25.79} & \textbf{55.60} & \textbf{26.30} \\
    \bottomrule
    \end{tabular}%
}

\resizebox{\LeftWidthC}{\TableHeightC}{\usebox{\leftboxC}}%
\hfill
\resizebox{\RightWidthC}{\TableHeightC}{\usebox{\rightboxC}}%

\end{table*}
}

\begin{abstract}

Flow Matching (FM) has achieved remarkable generative performance, yet it suffers from exposure bias due to discrepancies between training and inference. Existing mitigation strategies typically rely on static constraints or external heuristics. In this work, we propose that exposure bias itself inherently contains dynamic signals that can guide its own rectification. To leverage this, we introduce \textbf{\ourall{}} (\textbf{D}ir\textbf{E}ctional-\textbf{F}requency \textbf{A}daptive \textbf{R}ectification). This framework simulates the single-step inference process during training to identify exposure bias. It utilizes the directional and frequency adaptive feedback signals in bias itself to enhance the bias tolerance of the model. 
It consists of two key components: (1) \textbf{Anti-Drift Rectification (ADR)}. \ourdri{} treats inference-time drift as a signal to learn the direction to steer deviated states back toward the target. \ourdri{} endows the model with intrinsic active self-rectification capabilities; (2) \textbf{Frequency Compensation (FC)}. Empirically, we observe that accumulated bias often stems from a lack of low-frequency components in high-noise stages and exposure bias carries the missing frequency. \ourfre{} leverages the bias itself as a self-feedback weighting factor to reinforce the missing frequency components. Experiments on CIFAR-10, CelebA-64, and ImageNet-256/512 show that \ourall{} outperforms prior baselines and further demonstrates favorable scalability, compatibility, and inference robustness.
Code will be made available in \url{https://github.com/wuliwuliy/DEFAR}.
\end{abstract}
\sectionwithouttoc{Introduction}
\label{sec:intro}
% objective
Recently, Flow Matching (FM) \cite{building2023,flow2023,flowstraight2023} has emerged as a promising framework for generative modeling. By modeling continuous-time flows, FM offers a more direct and efficient training paradigm compared to traditional diffusion-based approaches \cite{ddpm2020,deep2015,generative2019}. Its strong empirical performance and theoretical simplicity have established FM as a foundational method across diverse generation tasks, including image, video, and audio generation \cite{SiT2024,flux2025,longcat2025,wan2025,sound2025,kimi2025,huangxuyue}.

\begin{wrapfigure}{r}{0.5\textwidth} 
    \centering
    \includegraphics[width=1\linewidth]{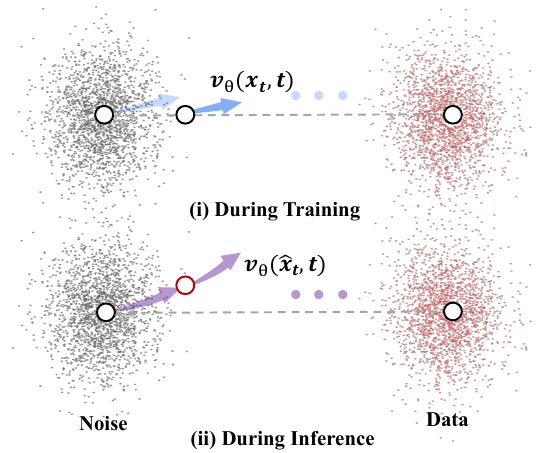}
    \caption{\textbf{Illustration of exposure bias in Flow Matching.} (i) During training, the model is conditioned on perturbed inputs sampled strictly along the ideal linear path (\textcolor[HTML]{3D83F5}{blue} arrows). (ii) During inference, the input suffers from accumulated errors generated by previous steps, causing training-inference mismatch (\textcolor[HTML]{6A4C7A}{purple} arrows).}
    \label{fig:intro_exposure_bias}
\end{wrapfigure}

However, FM models remain inherently vulnerable to exposure bias, a fundamental challenge stemming from the discrepancies between training and inference inputs. As illustrated in Fig. \ref{fig:intro_exposure_bias}, during training, the model is conditioned on perturbed input, namely a mixture of Gaussian noise and ground-truth data. The model effectively learns to predict the vector field without predicted error accumulation. 
In contrast, during multi-step inference, the model must exclusively rely on its own previous predictions.
This propagates and accumulates the potential predicted error at each inference step. 
This forces the model to operate in the unexplored data space compared to the training stage where its predictive capability significantly degrades, ultimately leading to substantial deviation from the target distribution.

While prior research has explored this issue within the DDPM framework \cite{error2024,frequency2025,alleviate2024,elucidating2024,exploiting2024}, existing solutions predominantly rely on static alignment strategies or exogenous noise injection. Some approaches introduce inference-time regularization without altering the training process \cite{alleviate2024,elucidating2024,frequency2025}, while others attempt to mitigate bias during training by simulating potentially imperfect inference. For instance, IP\cite{IP2023} injects fixed perturbations into training input data, and SS\cite{SS2022} directly replaces perturbed samples with predicted samples. These methods generally enforce a static alignment with the original ground truth or rely on externally fixed noise augmentation to improve robustness. Notably, \cite{multistep2024} proposes Multi-step Denoising Scheduled Sampling (MDSS) and Single-step Denoising Scheduled Sampling (SDSS), which focus on passively enhancing the model's robustness to input deviations. By exposing the model to perturbed states, MDSS encourages stability against accumulated errors but generally enforces a fixed alignment target regardless of the deviation's severity. In this work, we advance beyond passive robustness to establish a dynamic correction mechanism. We empower the model to learn adaptive correction signals that are modulated by the severity of exposure bias.

% In this work, we propose a dynamic correction paradigm that exploits the intrinsic self-correcting nature of exposure bias. Unlike static approaches, our method adaptively modulates the correction intensity based on the magnitude of the exposure bias, effectively creating a closed-loop feedback mechanism without requiring external noise injection.

% it can dynamically adjust the velocity field based on the magnitude and direction of the actual shift, enabling the model to adapt to varying degrees of exposure bias. 

% However, inputs with different prediction errors exhibit varying degrees of drift, as illustrated in Figure \ref{fig:exposure_bias_sample_dependent}. Thereby necessitating distinct levels of rectification. 
% Merely employing externally fixed or static correction strategies may lead to suboptimal performance. 
% This limitation leads us to a pivotal question: \textit{By simulating exposure bias during training, can we enable the model to autonomously learn adaptive rectification strategies tailored to the exposure bias on input?}

% They overlook a critical insight: the exposure bias itself is not merely noise to be suppressed from the outside, but contains endogenous dynamic signals that can be leveraged for real-time self-correction.

Specifically, we investigate the intrinsic properties of exposure bias from both directional and frequency perspectives.
\textbf{First}, we simulate the inference process during training to expose drift. 
After inference, the exposure bias is inherently carried in the inferred state. 
And the simulation enables us to explicitly characterize the deviation from the state to the target endpoint. 
Building on this, we introduce a directional regularizer. This term acts as a steering force, guiding the model to rectify the deviated state back towards the target endpoint. This allows us to actively adjust the rectification magnitude according to the severity of the exposure bias itself.
\textbf{Second}, identifying that exposure bias stems from prediction imperfection, we conduct a frequency analysis and observe a distinct deficiency in low-frequency components during the early denoising outputs from the model. 
Since these components encode critical structural information, their absence exacerbates error propagation. To quantify this, we introduce two metrics: Predicted Frequency Ratio (PFR), which measures the ratio of low-to-high frequency energy in prediction outputs, and Frequency Emphasis of Loss (FEL), which evaluates the relative focus of the loss function on different frequency bands. Intriguingly, we observe that exposure bias itself exhibits a complementary characteristic in relatively early high-noise timesteps: it naturally highlights neglected low-frequency regions. This motivates us to use the bias as a dynamic reweighting signal to mitigate itself.

Building on these insights, we propose \textbf{\ourall{}} (\textbf{D}ir\textbf{E}ctional-\textbf{F}requency \textbf{A}daptive \textbf{R}ectification), a flexible framework that leverages the dynamic nature of exposure bias itself to adaptively rectify FM models. \ourall{} comprises two core components: (1) Anti-Drift Rectification (ADR): Functioning as a regularizer to the original training objective, ADR constructs a direction that explicitly steers the model from the deviated state back towards the target data distribution, actively correcting the prediction drift based on the severity of exposure bias itself. 
(2) Frequency Compensation (FC): Applied to the original training objective, FC utilizes the exposure bias itself as a reweighting signal to enhance the learning of missing low-frequency components via a negative feedback loop. Our main contributions are summarized as follows:
\begin{itemize}
   \item We propose Anti-Drift Rectification (\ourdri{}), a method that actively provides directional correction, endowing flow matching with superior resistance to bias compared to static alignment methods.
   \item We identify the connection between exposure bias and missing low-frequency information, proposing Frequency Compensation (\ourfre{}) to dynamically reinforce structural learning by leveraging the bias signal itself.
   \item We integrate \ourdri{} and \ourfre{} into a unified framework, \ourall{}. Extensive experiments on CIFAR-10, ImageNet-256/512, and CelebA-64 demonstrate that \ourall{} significantly outperforms existing baselines, while exhibiting scalability, architectural compatibility, and inference robustness.
\end{itemize}

\sectionwithouttoc{Related Work}
\label{sec:Related_Work}
\textbf{Exposure Bias.} 
Exposure bias arises from the mismatch between training and inference inputs in sequential models \cite{NMT2020,closer2019,planner2023,sequence2015,scheduled2015,bridging2019}. Among early approaches, DAD \cite{data2014} combines ground truth and predicted tokens during training, and SS \cite{SS2022} samples biased inputs to better approximate inference.
In diffusion models, DDPM-IP \cite{IP2023} first formalizes exposure bias and perturbs training samples to simulate it. EP-DDPM \cite{error2024} studies error propagation and introduces a cumulative-error regularizer. AE-DDPM \cite{anti2025} mitigates exposure bias via prompt learning, and MCDO \cite{manifold2025} imposes manifold constraints. Inspired by distribution adaptation methods~\cite{dinglaodtbs,dinglaocvpr,zhangnips}, training-free techniques include a time-shift sampler (TS-DDPM \cite{alleviate2024}), noise scaling (ES-DDPM \cite{elucidating2024}), training-distribution leak signals exploitation\cite{exploiting2024} and frequency-domain analysis with wavelet-based regularization \cite{frequency2025}.
Among prior works, MDSS/SDSS~\cite{multistep2024} is most relevant to our approach. It simulates multi- or single-step inference trajectories during training but persists in aligning predictions with the static ground-truth target. Consequently, this strategy merely bolsters the model's passive robustness against input perturbations, failing to equip it with the capability to actively rectify accumulated errors. In contrast, we investigate the intrinsic signal properties of exposure bias in Flow Matching and introduce a novel paradigm that leverages the bias itself to drive active self-rectification.
% We instead align the prediction of model to a reconstructed target, enabling more effective correction of exposure bias.
\\
\textbf{Reweighting loss for emphasizing salient area alignment.} 
Several works explore reweighting strategies in loss functions to emphasize salient or informative regions, guiding models to focus on the most influential areas.
In decomposable or multi-stage generation, Decomposable Flow Matching \cite{improve2025} applies spatially varying masks to highlight critical regions on outputs.
In video generation, Proteus-ID \cite{proID2025} and MotiF \cite{motif2025} adopt motion-aware weighting to prioritize frequently moving regions. MotionCharacter \cite{motion2024} further improves learning efficiency by emphasizing motion-sensitive areas.
Beyond spatial analysis, Latent Wavelet Diffusion \cite{wavelet2025} performs frequency analysis and reweights loss to concentrate learning on frequency dominant components. Together, these methods demonstrate that adaptive loss reweighting can effectively direct model capacity toward salient regions, improving robustness and representation efficiency.
\sectionwithouttoc{Preliminaries}
\label{sec:preliminaries}
\subsectionwithouttoc{Flow Matching}
\label{sec:preli_flowmatching}
We adopt the standard Flow Matching formulation~\cite{flow2023}, where time $t \in [0, 1]$ drives the flow from the source noise $\pmb{\epsilon} \!\sim\! \mathcal{N}(\mathbf{0}, \mathbf{I})$ at $t=0$ to the target data $\mathbf{x}_* \!\sim\! p_{\text{data}}(\mathbf{x})$ at $t=1$.
A sample at time $t$ is constructed as
\begin{equation}
\label{eq:fm_path}
\mathbf{x}_t = a_t \mathbf{x}_* + b_t \pmb{\epsilon},
\end{equation}
where $a_t$ and $b_t$ are predefined schedules. 
The corresponding conditional velocity is given by the time derivative of the path:
$\mathbf{v}_t = a'_t \mathbf{x}_* + b'_t \pmb{\epsilon}$, where $a'_t = \frac{da_t}{dt}$ and $b'_t = \frac{db_t}{dt}$.
FM trains a neural network $\mathbf{v}_{\theta}(\mathbf{x}, t)$ to approximate this conditional velocity by minimizing
\begin{equation}
\label{eq:fm_loss}
\mathcal{L}_{\text{FM}}(\theta) := 
\mathbb{E}_{\mathbf{x}_*, \pmb{\epsilon}, t}
\!\left[
\big\|
\mathbf{v}_{\theta}(\mathbf{x}_t, t)
- (a'_t \mathbf{x}_* + b'_t \pmb{\epsilon})
\big\|^2
\right].
\end{equation}
This objective encourages the model to learn a time-dependent velocity field that consistently transports the intermediate distribution towards the true data distribution. For instance, to ensure the flow follows Optimal Transport (OT), we can employ the straight-line schedule $a_t=t$ and $b_t=1-t$, which simplifies the target velocity to a constant $\mathbf{x}_* - \pmb{\epsilon}$ along the linear path.

% \label{sec:preli_flowmatching}
% We adopt the standard Flow Matching formulation~\cite{flow2023}, where time $t \in [0, 1]$ drives the flow from the source noise $\pmb{\epsilon} \!\sim\! \mathcal{N}(\mathbf{0}, \mathbf{I})$ at $t=0$ to the target data $\mathbf{x}_* \!\sim\! p_{\text{data}}(\mathbf{x})$ at $t=1$.
% A sample at time $t$ is constructed as
% \begin{equation}
% \label{eq:fm_path}
% \mathbf{x}_t = a_t \mathbf{x}_* + b_t \pmb{\epsilon},
% \end{equation}
% where $a_t$ and $b_t$ are predefined schedules. 
% The corresponding velocity is
% $\mathbf{v}_t = \frac{da_t}{dt}\mathbf{x}_* + \frac{db_t}{dt}\pmb{\epsilon} = \mathbf{x}_* - \pmb{\epsilon}$,
% which remains constant along the linear path.
% FM trains a neural network $\mathbf{v}_{\theta}(\mathbf{x}, t)$ to approximate this conditional velocity by minimizing
% \begin{equation}
% \label{eq:fm_loss}
% \mathcal{L}_{\text{FM}}(\theta) := 
% \mathbb{E}_{\mathbf{x}_*, \pmb{\epsilon}, t}
% \!\left[
% \big\|
% \mathbf{v}_{\theta}(\mathbf{x}_t, t)
% - (\mathbf{x}_* - \pmb{\epsilon})
% \big\|^2
% \right].
% \end{equation}
% This objective encourages the model to learn a time-dependent velocity field that consistently transports intermediate distribution towards the true data distribution. To ensure the flow follows the Optimal Transport (OT), we employ the straight-line schedule $a_t=t$ and $b_t=1-t$.

\subsectionwithouttoc{Fourier Frequency Analysis}
\label{sec:preli_fourier}
The Fourier transform provides a way to represent spatial-domain signals in the frequency domain, decomposing them into sinusoidal components of different frequencies and amplitudes. Since the predicted velocity field $\mathbf{v} \in \mathbb{R}^{H \times W}$ shares a similar spatial structure with images, we can directly apply frequency-domain analysis to it. The 2D Discrete Fourier Transform (DFT)
of $\mathbf{v}$ is defined as:
\begin{equation}
\label{eq:dft}
\begin{split}
\mathbf{V}(u,v) = 
\sum_{x=0}^{H-1} \sum_{y=0}^{W-1}
\mathbf{v}(x, y)\, e^{-i 2\pi \left( \frac{ux}{H} + \frac{vy}{W} \right)},
\end{split}
\end{equation}
where $(u,v)$ denotes the frequency coordinates, and $i=\sqrt{-1}$ is the imaginary unit. The resulting $\mathbf{V}(u,v)$ consists of a real part $ R(\mathbf{V})$ and an imaginary part $ I(\mathbf{V})$, from which the amplitude and phase spectra can be derived as:
\begin{equation}
\label{eq:amp_phase}
\begin{split}
|\mathbf{V}(u,v)| &= \sqrt{ R(\mathbf{V})^2 + I(\mathbf{V})^2}, \\
\angle \mathbf{V}(u,v) &= \arctan\!\left(\frac{I(\mathbf{V})}{R(\mathbf{V})}\right).
\end{split}
\end{equation}

The inverse Fourier transform reconstructs the spatial-domain signal from its frequency representation:
\begin{equation}
\label{eq:idft}
\begin{split}
\mathbf{v}(x,y) = 
\frac{1}{HW}\!
\sum_{u=0}^{H-1}\sum_{v=0}^{W-1}
\mathbf{V}(u,v)\, e^{i 2\pi \left( \frac{ux}{H} + \frac{vy}{W} \right)}.
\end{split}
\end{equation}

In practice, we adopt the Fast Fourier Transform (FFT) and its Inverse (IFFT) for efficient computation.

\subsection{Exposure Bias Problem Formulation}
\label{sec:problem_formulation}
Exposure bias is a well-documented phenomenon in generative sequence modeling~\cite{sequence2015,scheduled2015} and has been recently investigated within diffusion-based frameworks~\cite{IP2023, multistep2024}. In the context of Flow Matching (FM), this issue manifests as a critical discrepancy between the training and inference phases. Specifically, during training, the model $\mathbf{v}_{\theta}(\mathbf{x}_t, t)$ relies on inputs $\mathbf{x}_t$ derived from a linear interpolation between the true data distribution $\mathbf{x}_*$ and Gaussian noise $\pmb{\epsilon}$ (Eq.~\ref{eq:fm_path}). Conversely, during inference, the input $\hat{\mathbf{x}}_t$ is generated recursively from previous predictions, inevitably accumulating errors. This mismatch induces a distribution drift between the training state $\mathbf{v}_{\theta}(\mathbf{x}_t, t)$ and the inference state $\mathbf{v}_{\theta}(\hat{\mathbf{x}}_t, t)$, which propagates through timesteps and progressively degrades sample quality.

\begin{figure*}[t]
    \centering
    \includegraphics[width=0.85\linewidth,height=0.16\textheight]{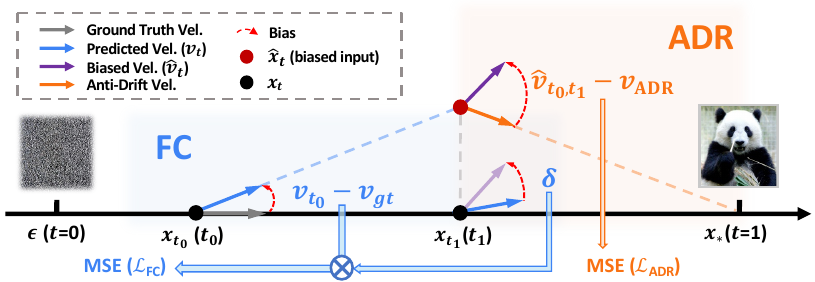}
    \caption{\textbf{Overview of \ourall{}.}
    (a) \textcolor[HTML]{F97112}{\textbf{Anti-Drift Rectification}}: Introduces a learning target that actively guides the model from the drift-affected distribution back toward the data distribution. 
    (b) \textcolor[HTML]{3D83F5}{\textbf{Frequency Compensation}}: Reweights the original objective using exposure bias as a negative-feedback signal to mitigate low-frequency deficiency at relatively high-noise timesteps (e.g., $t_0$). Together, \ourall{} adaptively rectifies exposure bias across both directional and frequency dimensions by leveraging the bias signal itself.
    }
    \label{fig:overview}
\end{figure*}

To explicitly model this effect during the training phase, we simulate the generation process to capture the resulting drift. As illustrated in Fig.~\ref{fig:overview}, we incorporate a one-step inference simulation in the training loop. At time $t_0$, given the ideal perturbed input $\mathbf{x}_{t_0}$, the model predicts the velocity $\mathbf{v}_{\theta}(\mathbf{x}_{t_0}, t_0)$. Due to unavoidable prediction errors, the estimated velocity naturally deviates from the ground-truth direction. Proceeding to the next timestep $t_1$, the resulting state is computed as: 
\begin{equation}
    \hat{\mathbf{x}}_{t_0,t_1} = \mathbf{x}_{t_0} + (t_1 - t_0) \cdot \mathbf{v}_{\theta}(\mathbf{x}_{t_0}, t_0).
\end{equation}
This estimated state carries the prediction drift. Consequently, the subsequent velocity $\mathbf{v}_{\theta}(\hat{\mathbf{x}}_{t_0,t_1}, t_1)$ exacerbates the deviation from the previous step.

Based on this simulation, we operationalize exposure bias at a single timestep to guide our training. We formulate the bias for any timestep interval $t_0, t_1$ as:
\begin{equation}
\label{eq:exposure_bias}
    \pmb{\delta}_{t_0,t_1} = \mathbf{v}_{\theta}(\hat{\mathbf{x}}_{t_0,t_1}, t_1) - \mathbf{v}_{\theta}(\mathbf{x}_{t_1}, t_1).
\end{equation}
This formulation quantifies the variation in predicted velocity caused solely by replacing the ideal input $\mathbf{x}_{t_1}$ with the drift-affected input $\hat{\mathbf{x}}_{t_0,t_1}$, serving as a tractable proxy for the actual inference discrepancy.

\sectionwithouttoc{DirEctional-Frequency Adaptive Rectification}
\label{sec:method}
This section presents our method, \textbf{\ourall{}}, a framework designed to adaptively mitigate exposure bias by leveraging the intrinsic directional and frequency signals of the bias itself. As illustrated in Fig.~\ref{fig:overview}, our approach comprises two core components: \textit{Anti-Drift Rectification} (\ourdri{}) and \textit{Frequency Compensation} (\ourfre{}), detailed in Sec.~\ref{subsec:adr} and Sec.\ref{subsec:frequency_compensate} respectively, with Sec.\ref{subsec:Frequency_Metrics} providing the metrics used for the subsequent frequency analysis. \ourdri{} operates in the directional dimension, actively learning a restorative direction to steer drifted states back toward the target endpoint based on the severity of the bias. 
In parallel, \ourfre{} functions in the frequency dimension, leveraging the frequency signature of the bias as a self-feedback cue to adaptively compensate for deficient low-frequency components during high-noise timesteps.
By integrating them into the unified \textbf{\ourall{}} framework, these two mechanisms complement each other, bolstering the bias tolerance of the model to training-inference mismatch from both directional and frequency dimensions. The learning objective combines the two mechanisms as follows:
\begin{equation}
\label{eq:all_methods}
\mathcal{L} := \beta_1 \mathcal{L}_{\text{\ourdri{}}} + \beta_2 \mathcal{L}_{\text{\ourfre{}}},
\end{equation}
where $\beta_1$ and $\beta_2$ are weighting coefficients for $\ourdri{}$ and $\ourfre{}$, respectively. $\mathcal{L}_{\text{\ourfre{}}} $ represents the $\mathcal{L}_{\text{FM}}$ reweighted by exposure bias signal. \ourall{} thus offers an effective solution to alleviate exposure bias by exploiting the informative signals inherent in the bias itself. The training procedure is summarized in Algorithm~\ref{alg:ours_algorithm}.

\subsectionwithouttoc{Anti-Drift Rectification (\ourdri{})}
\label{subsec:adr}
As formulated in Sec.~\ref{sec:problem_formulation}, exposure bias emerges from the recursive accumulation of prediction errors, integrating drift into $\hat{\mathbf{x}}_{t_0,t_1}$. When existing Flow Matching frameworks process such biased states, they typically impose a static optimization objective~\cite{multistep2024}, which lacks sensitivity to the dynamically varying severities of prediction drift. To empower the model with active resistance, we introduce Anti-Drift Rectification (\ourdri{}), a mechanism that turns the exposure bias into an endogenous control signal to mitigate itself.

Our fundamental motivation is grounded in the invariance of the target data distribution. Regardless of the severity of the accumulated exposure bias, the ground-truth endpoint $\mathbf{x}_*$ remains deterministic. This invariance allows us to establish a dynamic anchor: the anti-drift target velocity $\mathbf{v}_{\text{ADR}} = a'_{t_1} \mathbf{x}_* + b'_{t_1} \hat{\mathbf{x}}_{t_0,t_1}$. This vector acts as a restorative trajectory, originating strictly from the current biased state and guiding toward the target distribution. By leveraging this anchor, the model is empowered to actively recognize the precise corrective direction required to navigate back to the target manifold, adapting seamlessly to varying degrees of exposure bias. 

To enable the model to learn this rectifying direction, we introduce \ourdri{} as a regularization term appended to the standard FM objective (which will be further upgraded to our \ourfre{} objective in Sec.~\ref{subsec:frequency_compensate}). Notably, when no exposure bias occurs, $\mathbf{v}_{\text{ADR}}$ naturally degenerates to align parallel to $a'_{t_1} \mathbf{x}_* + b'_{t_1} \pmb{\epsilon}$. This guarantees consistency with the original probability flow, ensuring that the regularization only actively penalizes trajectory drift without perturbing optimal predictions. The \ourdri{} regularization term is formulated as follows:
\begin{equation}
 \label{eq:adr_loss}
 \begin{aligned}
 \mathcal{L}_{\text{\ourdri{}}} &= \mathbb{E}_{\mathbf{x}_*, \hat{\mathbf{x}}_{t_0,t_1}, t_0,t_1} \left[ \left\| \frac{\mathbf{v}_{\theta}(\hat{\mathbf{x}}_{t_0,t_1}, t_1)}{\|\mathbf{v}_{\theta}(\hat{\mathbf{x}}_{t_0,t_1}, t_1)\|_2} - \frac{a'_{t_1} \mathbf{x}_* + b'_{t_1} \hat{\mathbf{x}}_{t_0,t_1}}{\|a'_{t_1} \mathbf{x}_* + b'_{t_1} \hat{\mathbf{x}}_{t_0,t_1}\|_2} \right\|^2 \right].
 \end{aligned}
\end{equation}

This design serves as a dynamic self-rectifying mechanism: the regularization gradient is intrinsically proportional to the angle of deviation. As the exposure bias exacerbates the drift, the angular discrepancy widens, automatically amplifying the rectification signal. Through this dynamic modulation, \ourdri{} forces the model to actively learn a restorative direction, thereby alleviating sampling error and dismantling the cycle of recursive error propagation, thus enabling the model to leverage the exposure bias signal to mitigate the bias itself.

\subsectionwithouttoc{Frequency Analysis Metrics}
\label{subsec:Frequency_Metrics} 
Recent studies~\cite{alleviate2024,frequency2025,elucidating2024} attribute exposure bias to the distributional mismatch between training and inference predictions.
While \ourdri{} effectively rectifies the \textit{geometric direction} of this drift, the \textit{internal structure} of exposure bias, specifically its frequency characteristics, remains underexplored.
In this section, we analyze this pixel-level frequency mismatch and propose a fine-grained compensation strategy derived from the exposure bias itself. 

Inspired by the frequency analyses in~\cite{lin2023revisiting, tang2022rethinking}, we introduce two metrics to quantify the frequency distribution of velocity and loss for further analysis: the Predicted Frequency Ratio (PFR), which measures the ratio of low-to-high frequency energy in prediction, and Frequency Emphasis of Loss (FEL), which evaluates the relative focus of the loss function on different frequency bands.
% the Low-Frequency Dominant Ratio on Velocity ($R_\text{LFDV}$), which quantifies the dominance of low-frequency energy in velocity maps, and the Low-Frequency Dominant Ratio on Loss ($R_\text{LFDL}$), which measures the relative emphasis of the loss on low-frequency components over high-frequency ones with respect to the ground-truth velocity maps.

% We conduct a frequency-domain analysis on 
% for each sample and each channel, 
We first introduce the computation of PFR.
Specifically, given the predicted velocity $\mathbf{v} \in \mathbb{R}^{H \times W}$ produced by the FM model, we perform a 2D FFT on it:
\begin{equation}
\label{eq:fft}
\begin{split}
\mathbf{V}(u,v) = \text{FFT}(\mathbf{v}(i,j)).
\end{split}
\end{equation}

To investigate the proportion of low- and high-frequency bands, we apply Low-Pass Filter ($\mathbf{F}_{LP}$) and High-Pass Filter ($\mathbf{F}_{HP}$) defined as:
\begin{equation}
\label{eq:filters}
\begin{split}
\mathbf{F}_{\text{LP}}(u,v) = \mathbb{I}\left[ D(u,v) \leq D_{\text{cutoff}} \right], \quad
\mathbf{F}_{\text{HP}}(u,v) = 
\mathbf{1} - \mathbf{F}_{\text{LP}}(u,v),
\end{split}
\end{equation}
where $D(u,v) = \sqrt{u^2 + v^2}$ is the frequency magnitude and $D_{\text{cutoff}}$ is the cutoff frequency. We compute the energy ratio between the low- and high-frequency components and define PFR as follows:
\begin{equation}
\label{eq:energy_ratio}
\begin{split}
\text{PFR} := 
\frac{\sum_{u,v} ||\mathbf{V}(u,v)||^2 \cdot \mathbf{F}_{\text{LP}}(u,v)}
     {\sum_{u,v} ||\mathbf{V}(u,v)||^2 \cdot \mathbf{F}_{\text{HP}}(u,v)}.
\end{split}
\end{equation}

We then define the FEL. Here, we consider the target velocity map $\mathbf{v}_{\text{target}} \in \mathbb{R}^{H \times W}$ (typically the standard FM target velocity) and the MSE loss map $\pmb{\mathcal{L}} \in \mathbb{R}^{H \times W}$.
% where $\pmb{\mathcal{L}}$ denotes the per-pixel reconstruction loss before spatial averaging.
To compute it, we divide $\mathbf{v}_{\text{target}}$ into high- and low-frequency regions. The magnitude of the loss map $\pmb{\mathcal{L}}$ in each region reflects the relative emphasis of loss. Thus, we separately sum the loss values within the low- and high-frequency regions to quantify how the loss distributes its emphasis across different frequency components. This metric further serves as an indicator of the capability of loss to compensate for frequency discrepancies.

We first apply Eq.~\ref{eq:fft} and \ref{eq:filters} to perform the FFT of $\mathbf{v}_{\text{target}}$ and compute the corresponding low- and high-frequency masks in the frequency domain. These masks are then used to separate $\mathbf{v}_{\text{target}}$ into its low- and high-frequency components, which are subsequently transformed back to the spatial domain via IFFT. Finally, we compute the salient low- and high-frequency region masks in the spatial domain by thresholding the corresponding values according to their percentile distributions.
\begin{equation}
\label{eq:ifft_masks}
\begin{split}
\mathbf{v}_{\text{low}} &= \text{IFFT}(\mathbf{V} \cdot \mathbf{F}_{\text{LP}}), \quad
\mathbf{v}_{\text{high}} = \text{IFFT}(\mathbf{V} \cdot \mathbf{F}_{\text{HP}}),\\
\mathbf{M}_{\text{LFR}}(i,j) &= \mathbb{I}\big[\mathbf{v}_{\text{low}} > p(\mathbf{v}_{\text{low}})\big], \quad 
\mathbf{M}_{\text{HFR}}(i,j) = \mathbb{I}\big[\mathbf{v}_{\text{high}} > p(\mathbf{v}_{\text{high}})\big],
\end{split}
\end{equation}
where $p(\cdot)$ denotes a percentile threshold function, $\mathbf{v}_{\text{low}}$ and $\mathbf{v}_{\text{high}}$ represent the low- and high-frequency components of $\mathbf{v}_{\text{target}}$. The resulting masks, $\mathbf{M}_{\text{LFR}}$ and $\mathbf{M}_{\text{HFR}}$, indicate the low- and high-frequency region masks, respectively. 
% The separated regions are visualized in Fig.~\ref{fig:vt_visualization}.
Finally, we compute FEL as follows:
\begin{equation}
\label{eq:R_LFDL}
\begin{aligned}
\text{FEL} &:= 
\frac{\sum \tilde{\pmb{\mathcal{L}}} \cdot \mathbf{M}_{\text{LFR}}}
     {\sum \tilde{\pmb{\mathcal{L}}} \cdot \mathbf{M}_{\text{HFR}}}, \quad
\text{where} \quad
\tilde{\pmb{\mathcal{L}}}^{(i,j)} =
\frac{\pmb{\mathcal{L}}^{(i,j)}}
     {\sum_{i=1}^{H} \sum_{j=1}^{W} \pmb{\mathcal{L}}^{(i,j)}}.
\end{aligned}
\end{equation}

The trend of FEL reveals how the loss distributes attention across low- and high- frequency regions during training.

\subsectionwithouttoc{Frequency Compensation (\ourfre{})}
\label{subsec:frequency_compensate}

\begin{figure}[t]
  \centering
  \includegraphics[width=1\linewidth]{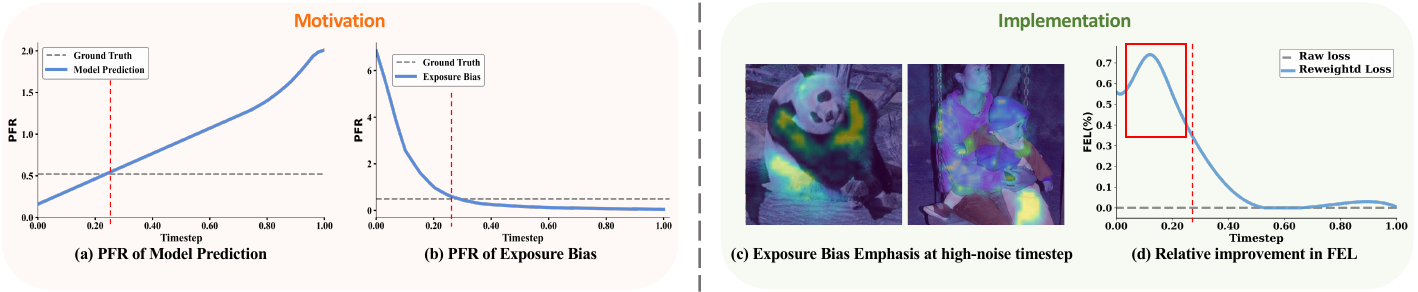}
  \caption{\textbf{Motivation and Verification of \ourfre{}.} (a) illustrates frequency trends derived from forward-perturbed inputs, while (b) reveals contrasting trends using inputs generated via single-step inference. In (b), for any starting timestep $t_0$, the exposure bias PFR is averaged over all subsequent sampled timesteps $t_1 \in (t_0, 1]$, capturing the cumulative impact across varying inference intervals. (c) visualizes the heatmap of exposure bias at high-noise timesteps, focusing on low-frequency semantics. (d) shows that the reweighted loss effectively emphasizes low-frequency regions during high-noise timesteps. The continuous time horizon $[0, 1]$ is discretized into $50$ uniform timesteps. The metrics are averaged across $50,000$ samples from ImageNet-256 on SiT-B/4.}
  \label{fig:FC_all}
\end{figure}

Leveraging the PFR metric proposed in Sec.~\ref{subsec:Frequency_Metrics}, we analyze the forward perturbation process and identify that prediction drift is primarily attributed to a frequency deficiency, which subsequently induces exposure bias.

\noindent\textbf{Exposure bias intrinsically compensates for the low-frequency deficiency during high-noise timesteps.} As visualized in Fig.~\ref{fig:FC_all} (a), the model prediction exhibits a distinct frequency shift over time. During the early high-noise steps ($t \to 0$), the prediction is characterized by a lack of low-frequency content (indicated by low PFR values) compared to the target $\mathbf{v}_{\text{target}}$. Conversely, Fig.~\ref{fig:FC_all} (b) reveals that the exposure bias follows an inverse trend, exhibiting high PFR values in this phase. This uncovers a complementary relationship: the exposure bias encapsulates the low-frequency structural components that the model inherently struggles to capture initially.

\noindent\textbf{Exposure bias serves as a dynamic and self-decaying corrective signal.} When approaching low-noise ($t \to 1$), the model progressively reconstructs sufficient low-frequency content. Concurrently, the low-frequency dominance within the exposure bias autonomously diminishes. This prevents excessive structural compensation, transforming the exposure bias from a static error into a dynamically weighted variable for self-correction.
\begin{algorithm}[htbp]
   \caption{\ourall{}: Training}
   \label{alg:ours_algorithm}
\begin{algorithmic}[1]

    \STATE {\bfseries Input:} model $f$ to predict velocity with parameters $\theta$, dataset $\mathcal{D}$, learning rate $\eta$, \ourdri{} loss weight $\beta_1$, \ourfre{} loss weight $\beta_2$, normalization coefficient $\alpha$, path interpolation coefficients $a_t, b_t$ (with derivatives $a'_t, b'_t$) and stability constant $\xi$.
    \STATE \textbf{While} $\theta$ not converged \textbf{do} 
    % Original indent: \STATE \hspace{1em} ...
    \STATE \quad $\pmb{\epsilon}\!\sim\!\mathcal{N}(0,I),\; \mathbf{x}_*\!\sim\! \mathcal{D},\; t_0, t_1 \in [0,1],\; \text{and } t_1 > t_0$

    \STATE \quad $\mathbf{x}_{t_0} \gets t_0\mathbf{x}_* + (1-t_0)\pmb{\epsilon}$, $\mathbf{x}_{t_1} \gets t_1\mathbf{x}_* + (1-t_1)\pmb{\epsilon}$ 
    
    \STATE \quad $\mathbf{v}_{t_0} \gets f_{\theta}(\mathbf{x}_{t_0},t_0), \mathbf{v}_{t_1} \gets \text{StopGradient}(f_{\theta}(\mathbf{x}_{t_1},t_1))$
    \STATE \quad $\hat{\mathbf{x}}_{t_0,t_1} \gets \mathbf{x}_{t_0} + (t_1 - t_0) \mathbf{v}_{t_0}$ \mycomment{Single-step inference simulation}
    \STATE \quad $\hat{\mathbf{v}}_{t_0,t_1}\gets f_{\theta}(\hat{\mathbf{x}}_{t_0,t_1},t_1)$
    \STATE \quad $\mathbf{v}_{\text{\ourdri{}}} \gets  a'_{t_1} \mathbf{x}_* + b'_{t_1} \hat{\mathbf{x}}_{t_0,t_1}$ \mycomment{ADR target velocity}
    \STATE \quad $\mathcal{L}_{\text{\ourdri{}}} \gets \mathbb{E} \left[ \left\| \frac{\hat{\mathbf{v}}_{t_0,t_1}}{\|\hat{\mathbf{v}}_{t_0,t_1}\|_2} - \frac{\mathbf{v}_{\text{\ourdri{}}}}{\|\mathbf{v}_{\text{\ourdri{}}}\|_2} \right\|^2 \right]$ \mycomment{ADR regularization loss}
    \STATE \quad $\pmb{\delta}_{t_0,t_1}\gets \hat{\mathbf{v}}_{t_0,t_1}-\mathbf{v}_{t_1}$
    \STATE \quad$ \mathbf{W}_{t_0,t_1}^{(i,j)} \gets 1 + \alpha \,
\frac{\|\pmb{\delta}_{t_0,t_1}^{(i,j)}\|^2}
     {\sum_{i=1}^{H} \sum_{j=1}^{W} \|\pmb{\delta}_{t_0,t_1}^{(i,j)}\|^2 + \xi}$
    \mycomment{Spatial frequency weights}
    \STATE \quad$\mathcal{L}_{\text{\ourfre{}}}\gets \mathbb{E} \!\left[
  \left\| 
    \mathbf{W}_{t_0,t_1} \cdot 
    \big( \mathbf{v}_{t_0} - (a'_{t_0} \mathbf{x}_* + b'_{t_0} \pmb{\epsilon}) \big)
  \right\|^2\right]$ \mycomment{FC-reweighted FM loss}
    \STATE \quad $\mathcal{L} \gets \beta_1\mathcal{L}_\text{\ourdri{}} + \beta_2\mathcal{L}_\text{\ourfre{}}$
    \STATE \quad $\theta \gets \theta - \eta \nabla_{\theta} \mathcal{L}$
\end{algorithmic}
\end{algorithm}\\
\noindent\textbf{The low-frequency content of exposure bias is semantically grounded in the target data.} To verify that this complementary signal is not noise, we visualize the spatial distribution of the exposure bias in Fig.~\ref{fig:FC_all} (c). The heatmaps confirm that during the high-noise timesteps, the bias is concentrated on the low-frequency structural regions of the original images (More samples in Appendix). 

These observations motivate our core insight: the dynamic frequency signatures of exposure bias constitute an endogenous feedback loop. This empowers the model to rectify low-frequency deficiencies using the bias signal, thereby reducing prediction errors and ultimately mitigating the bias itself.

Inspired by \cite{motif2025,wavelet2025,motion2024,proID2025}, which demonstrate that reweighting the loss can enhance saliency alignment, we design a negative-feedback weight mask for the original loss based on exposure bias. The weight is formally defined as:
\begin{equation}
\label{eq:W_exp_BC}
\begin{split}
\mathbf{W}_{t_0,t_1}^{(i,j)} = 1 + \alpha \,
\frac{\|\pmb{\delta}_{t_0,t_1}^{(i,j)}\|^2}
     {\sum_{i=1}^{H} \sum_{j=1}^{W} \|\pmb{\delta}_{t_0,t_1}^{(i,j)}\|^2 + \xi},
\end{split}
\end{equation}
where $\alpha$ is a scaling coefficient, and $\xi$ is a small constant introduced to ensure numerical stability.
We then reweight the FM learning objective as follows:
\begin{equation}
\label{eq:weighted_fm_loss}
\begin{split}
\mathcal{L}_{\text{\ourfre{}}} 
&:=\mathbb{E}_{\mathbf{x}_*,\,\pmb{\epsilon},t_0,t_1} \!\left[
  \left\| 
    \mathbf{W}_{t_0,t_1}\cdot 
    \big( \mathbf{v}_{\theta}(\mathbf{x}_{t_0},t_0) - (a'_{t_0} \mathbf{x}_* + b'_{t_0} \pmb{\epsilon}) \big)
  \right\|^2
\right].
\end{split}
\end{equation}

This weighting strategy adaptively adjusts the loss distribution according to the exposure bias, emphasizing the missing frequency components.

\noindent\textbf{Exposure bias reweighting prioritizes low-frequency learning during high-noise timesteps.} To quantitatively evaluate the impact of $\mathcal{L}_{\text{\ourfre{}}}$, we analyze the FEL. Prior studies \cite{freeinit24,fastinit2025, dinglaoacl,liangnips, identifying2024,bestnoise2025,luo2026beyond} indicate that the early generation phase, especially high-noise timesteps, is critical for establishing the global structure of generated samples, which directly influences the final generation quality. Consistent with these findings, as illustrated in Fig.~\ref{fig:FC_all} (d), the exposure-bias-reweighted loss exhibits a relative increase over the raw baseline, directing the learning focus toward low-frequency components during the high-noise phase ($t \to 0$). By utilizing the bias signal to compensate for frequency deficiencies, \ourfre{} translates the exposure bias into a corrective signal, thereby leveraging the bias to mitigate its underlying cause.
\sectionwithouttoc{Experiments}
\label{sec:experiments}
\subsectionwithouttoc{Experiment Setting} 
\textbf{Main Implementation.} 
We evaluate our methods on conditional image generation tasks on ImageNet-256/512 \cite{imagenet2009} and CIFAR-10 \cite{cifar10_2009}, as well as unconditional tasks on CIFAR-10 and CelebA-64 \cite{CelebA64_2015}. Following prior works \cite{mean2025,IP2023}, we generate 50K samples using 50 Number of Function Evaluations (NFE).
CIFAR-10 and CelebA-64 are processed in pixel space, while ImageNet-256 is encoded in the latent space of a pre-trained VAE-ft-EMA tokenizer~\cite{VAE2022}. All models are trained from scratch on Ascend 910B NPUs, except that REPA, DDT and OT-CFM experiments are conducted on A100 GPUs. Based on grid search results (detailed in Appendix), we set $\beta_1=10$, $\beta_2=1$ and $\alpha=1$ in all experiments.
During training, all variants of our methods randomly select a pair of timesteps $(t_0, t_1)$ with $t_1 > t_0$ in each iteration. For ablation studies, we adopt SiT-B/4 as the default backbone and $a_t=t,b_t=1-t$ unless stated otherwise.
For Discriminator Guidance (DG), we follow~\cite{kim2023refining} and train a shallow U-Net discriminator for the SiT-B/4 score model with an ADM classifier. For mini-batch OT coupling, we follow~\cite{tong2024improving} and report five-seed statistics.  More details in Appendix. % Added: DG and mini-batch OT experiment details.
\\
\textbf{Evaluation Metrics.} We report the training budget in NPU/GPU hours (GPU for REPA, DDT, and mini-batch OT coupling experiments), denoted as ``Hour''. We follow \cite{beatgans2021} for the computation of all evaluation metrics. Fréchet Inception Distance (FID) \cite{FID2017} is reported as our primary metric, alongside sFID \cite{sfid2021}, Inception Score (IS) \cite{IS2016}, Precision and Recall \cite{preandre2019}. Following prior works \cite{multistep2024,alleviate2024,beatgans2021}, we utilize the full training sets of CIFAR-10 and CelebA-64, and the standard testset of ImageNet, as our reference distributions. Details in Appendix.\\
\textbf{Frequency Analysis Details.} To maximize the separation between low- and high-frequency regions, the cutoff frequency is set to $\min(H, W) // 8$ in Eq.~\ref{eq:filters}, with 20\% and 25\% thresholds respectively in Eq.~\ref{eq:ifft_masks}.\\
\noindent\textbf{Baselines.} We compare our method with four baselines: SiT~\cite{SiT2024}, IP~\cite{IP2023}, SDSS~\cite{multistep2024}, and MDSS~\cite{multistep2024}. For IP, SDSS, and MDSS, which were originally developed under the DDPM framework, we adapt their core ideas to the flow matching setting. Specifically, IP perturbs the input with additional noise to simulate inference, SDSS aligns predictions with the static target during single-step inference in training, and MDSS performs multi-step inference (we use 4 steps as reported to achieve the best performance \cite{multistep2024}).
We further evaluate two complementary exposure-bias mitigation paradigms. DG \cite{kim2023refining} corrects generated states at inference time with external classifier and discriminator. Mini-batch OT coupling (OT-CFM) \cite{tong2024improving} changes the noise-data coupling during training to reduce transport-path variance. These two paradigms allow us to test whether \ourall{} remains effective when exposure bias is also mitigated by inference-time correction or coupling-based path straightening. % Added: describe DG and mini-batch OT baseline settings and why they are necessary.
\TabA
\TabAtwo
\subsectionwithouttoc{Quantitative Experiments}
To account for differences in computational environments, we reproduce the key baselines and report the results in Tab.~\ref{tab:major_conditional} and Tab.~\ref{tab:major_unconditional}. We highlight five primary findings: % Modified: add the complementary-baseline finding.
\textit{(i)} Under comparable training budgets, \ourall{} consistently improves upon the SiT baseline, reducing FID by \textbf{1.24/1.83} on ImageNet-256 without/with Classifier-Free Guidance (CFG), \textbf{2.08/1.63} on conditional CIFAR-10, and \textbf{1.48} and \textbf{1.13} on unconditional CIFAR-10 and CelebA-64, respectively. This confirms its effectiveness in enhancing generation quality. 
\textit{(ii)} \ourdri{} outperforms MDSS and SDSS~\cite{multistep2024}. Unlike these baselines, which rely on static alignment targets, \ourdri{} learns an active anti-drift target, validating the advantage of dynamic directional rectification. 
\textit{(iii)} Both \ourdri{} and \ourfre{} independently yield performance gains, while their integration within the unified \ourall{} framework achieves better results. 
\textit{(iv)} While prior methods passively mitigate exposure bias via fixed input perturbation (IP) or static target alignment (SDSS, MDSS), \ourall{} actively rectifies prediction drift and compensates for deficient frequency components, achieving consistent empirical gains over these approaches.
\textit{(v)} In Tab.~\ref{tab:major_unconditional} (right), the DG and mini-batch OT comparisons further verify complementarity. For DG, \ourall{} improves FID by \textbf{0.52} over DG alone, and combining \ourall{} with DG reduces FID by \textbf{1.89}, suggesting that \ourall{} strengthens the base model during training while DG provides complementary inference-time guidance. For mini-batch OT coupling, combining \ourall{} with OT-CFM achieves the best Avg. FID \textbf{3.960}{\scriptsize$\pm$0.034}, outperforming OT-CFM alone by \textbf{0.135} and \ourall{} alone by \textbf{0.096} in terms of the mean value. This suggests that OT-CFM reduces the transport-path difficulty and thereby provides a better basis for \ourall{} to perform accurate fine-grained rectification according to the magnitude of exposure bias, leading to further gains. % Modified: align DG/OT conclusions with parallel wording and table-readable five-seed statistics.
% Split Fig. 4 block. Move this block independently if needed.
\begin{figure*}[t]
    \centering
    \includegraphics[width=\linewidth,keepaspectratio]{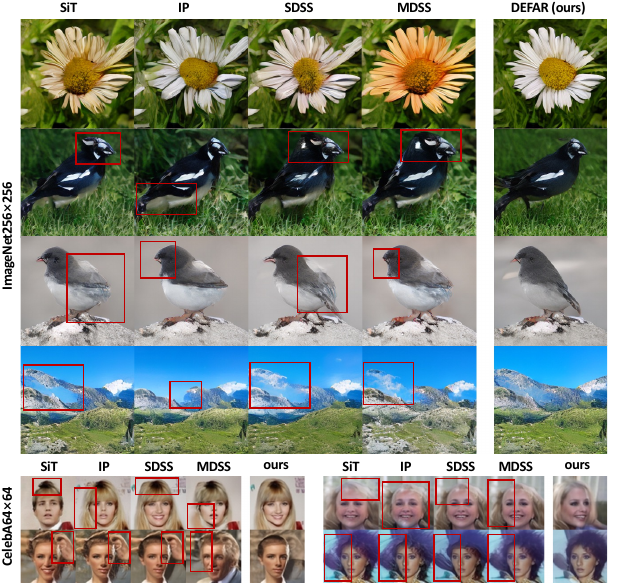}
    \caption{\textbf{Qualitative Comparison.} \ourall{} produces the most realistic samples on ImageNet-256 and CelebA-64. Red boxes highlight the blurriness or distorted details.}
    \label{fig:quality_comparison}
\end{figure*}
\subsectionwithouttoc{Qualitative Experiments}
All methods use the same random seed and 50 NFE for a fair comparison. As shown in Fig.~\ref{fig:quality_comparison}, baselines exhibit two clear visual artifacts: \textit{(i)} Rows 1, 2, and 5 exhibit severe structural disintegration and missing details. \textit{(ii)} Rows 3, 4, and 6 show blurred object-background boundaries. These issues arise from the exposure bias accumulating during high-noise timesteps, causing confusion between the primary object and the surrounding context. 
In contrast, \ourall{} generates images with more coherent structures and consistent textures.

\sectionwithouttoc{Discussion}

\begin{table*}[t]
    \centering
    \caption{\textbf{Scalability of \ourall{} on ImageNet-256.} 50-NFE generation (without/with CFG). \ourall{} consistently outperforms SiT across different scales.}
    \label{tab:scalability}
    \resizebox{\textwidth}{!}{%
        \begin{tabular}{l c c c c c c c c c}
        \toprule
        \multirow{2.5}{*}{\textbf{Method}} & \multirow{2.5}{*}{\textbf{Backbone}} &\multirow{2.5}{*}{\textbf{Params}} &\multirow{2.5}{*}{\textbf{Iters}} &\multirow{2.5}{*}{\textbf{Hour}} & \multicolumn{5}{c}{\textbf{ImageNet-256}}\\ 
        \cmidrule(lr){6-10}
        &&&&& FID$\downarrow$ & sFID$\downarrow$ & IS$\uparrow$& Pre.$\uparrow$& Rec.$\uparrow$\\ 
        \midrule
        SiT \cite{SiT2024} & SiT-B/4 & 131M& 500k & 166 & 61.64/31.64 & 12.20/8.44 & 24.67/55.65  & \textbf{0.387}/\underline{0.542} & 0.581/0.524\\
        \ourall{} & SiT-B/4 & 131M &250k& 144 & \underline{60.40}/\underline{29.81} & \textbf{9.05}/\textbf{7.35} & \underline{26.21}/\underline{58.81} & 0.375/0.531 & \underline{0.586}/\textbf{0.532}\\
        &&&&&\textcolor[HTML]{CC0000}{-1.24}/\textcolor[HTML]{CC0000}{-1.83} &\textcolor[HTML]{CC0000}{-3.15}/\textcolor[HTML]{CC0000}{-1.09}&\textcolor[HTML]{CC0000}{+1.54}/\textcolor[HTML]{CC0000}{+3.16}&\textcolor[HTML]{009900}{-0.012}/\textcolor[HTML]{009900}{-0.011}&\textcolor[HTML]{CC0000}{+0.005}/\textcolor[HTML]{CC0000}{+0.008}\\
        \ourall{} & SiT-B/4 & 131M &500k & 289 & \textbf{56.39}/\textbf{28.34}  & \underline{9.50}/\underline{7.84} & \textbf{26.62}/\textbf{58.94} & \underline{0.386}/\textbf{0.551} & \textbf{0.594}/\underline{0.524}\\
        \midrule
        SiT \cite{SiT2024} & SiT-M/2 & 308M & 700k & 311 & 30.57/8.23 & 10.90/6.13 & 58.66/154.71 & \underline{0.547}/0.750 & \underline{0.654}/\textbf{0.531}\\
        \ourall{} & SiT-M/2 & 308M & 350k & 330 & \underline{29.46}/\underline{7.85} & \textbf{5.60}/\textbf{4.64} & \underline{59.01}/\underline{156.27} & \textbf{0.549}/\underline{0.751} & 0.639/0.518\\
        &&&&&\textcolor[HTML]{CC0000}{-1.11}/\textcolor[HTML]{CC0000}{-0.38} &\textcolor[HTML]{CC0000}{-5.30}/\textcolor[HTML]{CC0000}{-1.49}&\textcolor[HTML]{CC0000}{+0.35}/\textcolor[HTML]{CC0000}{+1.56}&\textcolor[HTML]{CC0000}{+0.002}/\textcolor[HTML]{CC0000}{+0.001}&\textcolor[HTML]{009900}{-0.015}/\textcolor[HTML]{009900}{-0.013}\\
        \ourall{} & SiT-M/2 & 308M & 700k & 661 & \textbf{27.58}/\textbf{7.30} & \underline{7.35}/\underline{4.96} & \textbf{62.09}/\textbf{159.76} & 0.546/\textbf{0.752} & \textbf{0.658}/\textbf{0.531}\\
        \midrule
        SiT \cite{SiT2024} & SiT-XL/2 & 675M & 700k & 2260 & 25.85/5.88 & 7.63/\underline{4.73} & 68.93/183.58 & 0.570/0.762 & \textbf{0.659}/\textbf{0.541}\\
        \ourall{} & SiT-XL/2 & 675M & 350k & 2300 & \underline{24.93}/\underline{5.74} & \textbf{5.71}/\textbf{4.71} & \underline{71.63}/\underline{188.75} & \underline{0.580}/\underline{0.775} & \underline{0.651}/\underline{0.532}\\
        &&&&&\textcolor[HTML]{CC0000}{-0.92}/\textcolor[HTML]{CC0000}{-0.14} &\textcolor[HTML]{CC0000}{-1.92}/\textcolor[HTML]{CC0000}{-0.02}&\textcolor[HTML]{CC0000}{+2.70}/\textcolor[HTML]{CC0000}{+5.17}&\textcolor[HTML]{CC0000}{+0.010}/\textcolor[HTML]{CC0000}{+0.013}&\textcolor[HTML]{009900}{-0.008}/\textcolor[HTML]{009900}{-0.009}\\
        \ourall{} & SiT-XL/2 & 675M &700k & 4640 & \textbf{18.45}/\textbf{4.14} & \underline{6.34}/4.95 & \textbf{77.80}/\textbf{199.59} & \textbf{0.621}/\textbf{0.810} & 0.644/0.529\\
        \bottomrule
        \end{tabular}%
    }
\end{table*}

\TabC
% \vspace{-25pt}
\begin{figure*}[t]
    \centering
    \begin{adjustbox}{valign=t}
    \begin{minipage}{0.41\textwidth}
        \centering
        \includegraphics[width=0.90\linewidth]{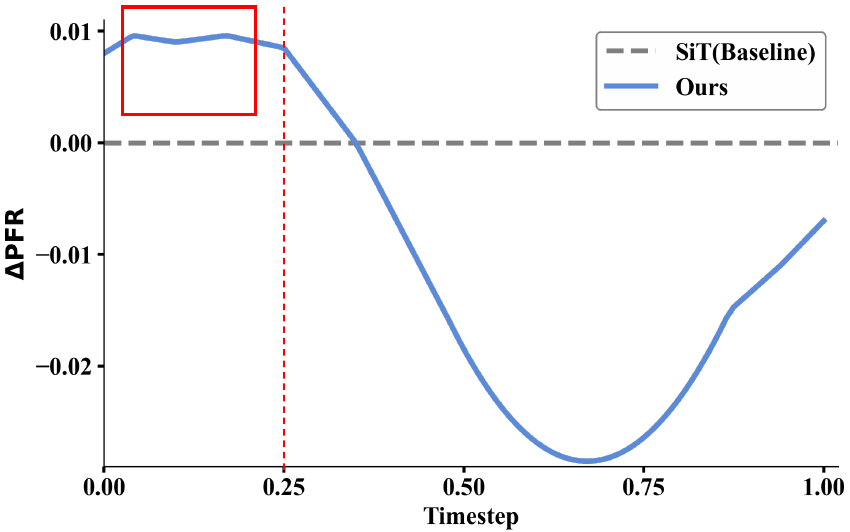}
        \captionof{figure}{\textbf{Low-frequency Restoration.} After comparable training, \ourall{} can compensate for missing low-frequency components during high-noise timesteps (\textcolor[HTML]{CC0000}{Red} box).}
        \label{fig:frequency_analysis_after_train}
    \end{minipage}
    \end{adjustbox}\hfill
    \begin{adjustbox}{valign=t}
    \begin{minipage}{0.55\textwidth}
        \centering
        {\captionsetup{type=table}%
        \captionof{table}{\textbf{Generalization across path interpolants and sampling methods.} FID on unconditional CIFAR-10.}
        \label{tab:different_interpolate}}
        \renewcommand{\arraystretch}{0.98}
        \smallskip
        \resizebox{0.93\linewidth}{!}{% 
        \begin{tabular}{c c c cc}
            \toprule
            \textbf{Interpolant} &\textbf{Model} &\textbf{Hour} & \textbf{ODE} & \textbf{SDE}\\ 
            \midrule
            Linear & SiT-B/4 &12  &17.41 & 16.81\\
            Linear & \ourall{}-B/4 &11 &\textbf{15.93} & \textbf{15.31} \\
            \midrule
            SBDM-VP & SiT-B/4 & 12 & 19.89 & 18.89 \\
            SBDM-VP & \ourall{}-B/4 & 11& \textbf{18.33} & \textbf{17.35}\\
            \midrule
            GVP & SiT-B/4 &12 & 17.31 & 16.42 \\
            GVP & \ourall{}-B/4 & 11&\textbf{15.84} & \textbf{14.91} \\
            \bottomrule
        \end{tabular}%
        }
    \end{minipage}
    \end{adjustbox}
\end{figure*}
In this section, we conduct a multifaceted evaluation of \ourall{}, assessing its scalability across model sizes, frequency restoration capability, and generalizability to diverse architectures and inference configurations.\\
\noindent\textbf{Scalability across Model Sizes.} Tab.~\ref{tab:scalability} reports FID alongside four metrics of \ourall{} across different model scales (B/4, M/2 and XL/2). \ourall{} consistently outperforms the SiT models and exhibits scalability in generative quality.\\
% \vspace{-12pt}
\noindent\textbf{Frequency Restoration Analysis.} As shown in Fig.~\ref{fig:frequency_analysis_after_train}, we compare the PFR of predicted velocities between \ourall{} and the SiT baseline when given forward-perturbed inputs.
\ourall{} effectively recovers low-frequency components during high-noise timesteps, alleviating the low-frequency deficiency.

\noindent\textbf{Universality across Architectures.} We integrate our method into two strong and distinct diffusion transformers: REPA~\cite{repa2025} (featuring representation alignment) and DDT~\cite{ddt2025} (featuring structural decoupling). We fine-tune their default REPA-XL/2 and DDT-XL/2 backbones on ImageNet-512 from the official ImageNet-256 pre-trained checkpoints. This demonstrates that our method seamlessly adapts to these advanced paradigms. 
As shown in Tab.~\ref{tab:compatibility_and_nfe} (left), under a comparable training budget, our approach consistently achieves superior generation quality.\\
% SBDM-VP (Variance Preserving), and GVP (Generalized Variance Preserving)
\noindent\textbf{Adaptability to Path Interpolants and Sampling Methods.} As shown in Tab.~\ref{tab:different_interpolate}, we examine three interpolant types: Linear, SBDM-VP, and GVP~\cite{SiT2024}. The results indicate that \ourall{} is adaptable to the choice of the forward process. Under both Ordinary Differential Equation (ODE) and Stochastic Differential Equation (SDE) sampling methods, \ourall{} yields performance gains.\\
\noindent\textbf{Robustness to Inference Steps.} 
Potential prediction bias can accumulate during inference. Increasing inference steps often amplifies exposure bias. To assess this effect, we evaluate all methods under various step settings. \ourall{} consistently achieves the lowest FID across nearly all settings, maintaining stable performance as the sampling step increases in Tab.~\ref{tab:compatibility_and_nfe} (right).
% \vspace{-2pt}
\sectionwithouttoc{Conclusion}
% \vspace{-1pt}
This work introduces \ourall{}, a framework designed to adaptively mitigate exposure bias by exploiting its intrinsic properties rather than relying on passive robustness. We demonstrate that exposure bias serves as a valuable signal, providing both directional guidance and frequency weighting cues. By integrating Anti-Drift Rectification (ADR) and Frequency Compensation (FC), \ourall{} empowers the model to perform dynamic self-rectification based on its own deviations. Extensive experiments verify the superior performance, scalability, compatibility, and inference robustness of \ourall{} across various benchmarks.

\section*{Acknowledgements}
The research of Shao-Lun Huang is supported in part by National Key R\&D Program of China under Grant 2021YFA0715202, the National Natural Science Foundation of China under Grants 62571296 and Huawei.

\bibliographystyle{splncs04}
\bibliography{main}

\clearpage
\appendix
% !TEX root = ../main_arxiv.tex
% \onecolumn
% \clearpage
\section{Notations}
\label{app:Notations}
% \onecolumn
\begingroup
    \setlength{\tabcolsep}{0pt}
    \renewcommand{\arraystretch}{1.2}
    \begin{center}
    \begin{adjustbox}{width=\textwidth}
  \begin{tabular}{lcr}
    \textbf{Symbol} & \textbf{Description} \\
    $\mathbf{x}_*$ & Target data\\
    $\mathbf{x}_{t}$ & The forward linear interpolation between data and Gaussian noise at timestep $t$  \\
    $\hat{\mathbf{x}}_{t_0,t_1}$ & The reverse predicted biased sample from timestep $t_0$ to timestep $t_1$\\
    $\hat{\mathbf{x}}_*$ & The final predicted biased endpoint\\
    $\pmb{\epsilon}$ & Gaussian noise\\
    $\mathbf{v}_t$ & The predicted velocity from the model taking the forward perturbed $\mathbf{x}_{t}$ as input \\

    $\mathbf{v}_{\text{target}}$ &  The FM target velocity, typically $\mathbf{x}_{*}-\pmb{\epsilon}$ \\
    $\hat{\mathbf{v}}_{t_0,t_1}$ & The predicted velocity from the model taking the reverse-predicted biased input $\hat{\mathbf{x}}_{t_0,t_1}$ as input\\

    $\mathbf{v}_{\text{ADR}}$ & The reconstructed anti-drift rectification target\\

    $\mathbf{V}$ & The frequency-domain representation of the velocity map\\

    $\mathbf{v}_{\text{low} / \text{high}}$ & The low- / high-frequency components on velocity map after IFFT transforming\\    
    $\mathbf{M}_{\text{LFR}/\text{HFR}}$ & The dominant low- / high-frequency mask of $\mathbf{v}_{\text{target}}$\\

    $t$ & The timestep and  $t \in [0,1] $ \\

    $ t_{0/1}$ & The starting / ending timestep of the single-step training-time inference process, with $t_1 > t_0$\\
    
    $\theta$ & Model parameters \\
    $f_{\theta}(\cdot)$ & The velocity prediction function parameterized by $\theta$
    \\
    $p(\cdot)$ & The distribution function\\
    $\mathcal{L}$ & The overall loss\\
    $\mathcal{L}_{\text{ADR}}$ & The anti-drift rectification loss\\

    $\mathcal{L}_{\text{FC}}$ & The frequency compensation loss\\

    $\pmb{\mathcal{L}}$ & The loss map with shape $H \times W$\\

    $\tilde{\pmb{\mathcal{L}}}$ & The normalized loss map of $\pmb{\mathcal{L}}$\\
    
    $\mathcal{N}$ & Normal distribution\\
    
    $\pmb{\delta}_{t_0,t_1} $ & The exposure bias during time interval from $t_0$ to $t_1$\\

    $\text{PFR}$ & The low-frequency components dominant ratio on velocity\\

    $\text{FEL}$ & The ratio in the loss that emphasizes the low-frequency components of $\mathbf{v}_{\text{target}}$\\

    $\text{DFT / FFT / IFFT}$ & Discrete Fourier Transform / Fast  Fourier Transform / Inverse Fast Fourier Transform\\
    $\mathbf{F}_{\text{LP}/\text{HP}}$ & The low-pass / high-pass filter\\
    $\mathbf{W}_{t_0,t_1}$ & The frequency-aware weight modulated by exposure bias \\
    $(u,v)$ & The coordinates of the frequency-domain signal
    \\
    $R$ & The real part of DFT result\\
    $I$ & The imaginary part of DFT result\\
    $\mathcal{D}$ & Training dataset \\
    $H \times W$ & The height and width shape of a sample\\
    $\alpha$ & The coefficient modulating the influence of exposure bias on the frequency-aware weight\\
    $\xi$ & The stability constant\\
    $\beta_1$ & The hyperparameter of \ourdri{} loss\\
    $\beta_2$ & The hyperparameter of \ourfre{} loss\\
    $\eta$ &  Learning rate\\
    $\pmb{\epsilon}'$ & The predicted Gaussian noise\\
    $a,b$ & The interpolation coefficients\\
    
  \end{tabular}
    \end{adjustbox}
    \end{center}
\endgroup

\clearpage
\section{Additional Ablation Studies}
This section presents further ablation studies on components of our proposed method, along with details regarding the implementation hyperparameters.

\subsectionwithouttoc{Effectiveness of Normalization for Anti-Drift Target}
\begin{table}[t]
\centering
\caption{\textbf{Directional normalization stabilizes training.} Results are reported using a SiT-B/4 model trained for 250k iterations on the CIFAR-10 conditional generation task with 50 NFEs, without Classifier-Free Guidance (CFG).}
\resizebox{0.8\textwidth}{!}{
\begin{tabular}{lccccc}
\toprule
\multirow{2.5}{*}{\textbf{Method}} &\multirow{2.5}{*}{\textbf{Normalization}}&\multirow{2.5}{*}{\textbf{Direction}} &\multirow{2.5}{*}{\textbf{Iteration}} & \multicolumn{2}{c}{\textbf{CIFAR-10 (cond)}}\\  
\cmidrule(lr){5-6}
 &&&& FID$\downarrow$ & IS$\uparrow$\\ 
\midrule
SiT & / & / &250k & 14.09 & 8.22\\
\ourall{}  & / & Anti-Drift & 250k & 317.31 & 1.38 \\
\ourall{}  & Rescale & Anti-Drift & 250k & 20.20 & 7.93 \\
\ourall{} & Unit Length & Directly Straight & 250k & 12.15 & 8.25\\
\ourall{} (ours)  & Unit Length & Anti-Drift & 120k  & \underline{12.01} & \underline{8.27} \\
\ourall{} (ours)  & Unit Length & Anti-Drift & 250k  & \textbf{11.22} & \textbf{8.31} \\
\bottomrule
\end{tabular}
}
\label{tab:anti-drift_target_normalization}
\end{table}

In this section, we further analyze why normalizing the anti-drift target is essential and provide empirical evidence to demonstrate its importance.

\noindent\textbf{Without normalization, the anti-drift target misguides the model toward noisy outputs.}
Removing both the flow-matching loss and the normalization, we force the model to directly predict the raw anti-drift target. Let $t=0$ denote the pure noise distribution and $t=1$ denote the data distribution. The generalized predicted velocity can be parameterized as:
\begin{equation}
\label{eq:raw_target_merged}
\mathbf{v}_\text{\ourdri{}} = a'_t\mathbf{x}_*+b'_t\hat{\mathbf{x}}_{t_0,t_1}.
\end{equation}
While Eq.~\ref{eq:raw_target_merged} formulates the target for a specific training interval from $t_0$ to $t_1$, the actual inference process entails a continuous integration. During inference, at any continuous timestep $t$, the model takes the intermediate state $\hat{\mathbf{x}}_t$ as input, which has accumulated drift from previous steps. Assuming the model perfectly learns this unnormalized target mapping and applies it at each continuous step $t$, the endpoint of the inference trajectory starting from pure noise $\mathbf{x}_0 = \pmb{\epsilon}$ is derived as:
\begin{equation}
\begin{aligned}
\label{eq:unnormalization_result_general}
\hat{\mathbf{x}}_* &= \pmb{\epsilon} + \int_0^1 (a'_t\mathbf{x}_* + b'_t\hat{\mathbf{x}}_t)\,dt \\
&= \pmb{\epsilon} + \left( \int_0^1 a'_t\,dt \right) \mathbf{x}_* + \int_0^1 b'_t\hat{\mathbf{x}}_t\,dt.
\end{aligned}
\end{equation}
In practice, the integral term $\int_0^1 b'_t\hat{\mathbf{x}}_t\,dt$ cannot completely cancel the initial noise $\pmb{\epsilon}$. This is because the intermediate state $\hat{\mathbf{x}}_t$ inherently deviates from the marginal distribution of the standard forward process due to the accumulated sampling drift. Consequently, the approximation error compounds along the denoising trajectory. Even with a perfectly learned target, the model lacks the capability to fully recover the ground truth sample $\mathbf{x}_*$, which explains the severe performance degradation when directly learning Eq.~\ref{eq:raw_target_merged}, as reported in the second row of Tab.~\ref{tab:anti-drift_target_normalization}.

\noindent\textbf{Rescaling the anti-drift target introduces training instability.}
Directly learning the anti-drift target is problematic because the distribution of \(\hat{\mathbf{x}}_{t_0,t_1}\) varies across different timestep intervals \(t_0,t_1\). A natural alternative is to rescale the anti-drift target so that its magnitude matches the standard flow matching objective \((a'_t\mathbf{x}_* + b'_t \pmb{\epsilon})\). Concretely, we divide the target by the effective time length:
\begin{equation}
\label{eq:rescaled_target}
    \mathbf{v}_{\text{Rescaled \ourdri{}}} = \frac{a'_t\mathbf{x}_*+b'_t\hat{\mathbf{x}}_{t_0,t_1}}{1-t_1}.
\end{equation}
This velocity alignment ensures that integrating the rescaled direction from \(0\) to \(1\) achieves the correct magnitude. 

However, during training, \(t_1\) is sampled uniformly over the full interval \([0,1]\). As \(t_1 \to 1\), the scaling factor \(1/(1-t_1)\) in Eq.~(\ref{eq:rescaled_target}) diverges, leading to exploding losses and numerical instability. This inherent instability is directly reflected in the degraded generation results shown in the third row of Tab.~\ref{tab:anti-drift_target_normalization}.

\noindent\textbf{Normalization ensures stable directional guidance for anti-drift rectification.}
To avoid numerical instability while preserving crucial directional information, we explicitly normalize both the predicted velocity and the anti-drift target to unit length. 
The final directional regularization term added to the flow-matching objective is defined as:
\begin{equation}
 \label{eq_appdix:adr_loss}
 \begin{aligned}
 \mathcal{L}_{\text{\ourdri{}}} &= \mathbb{E}_{\mathbf{x}_*, \hat{\mathbf{x}}_{t_0,t_1}, t_0,t_1} \left[ \left\| \frac{\mathbf{v}_{\theta}(\hat{\mathbf{x}}_{t_0,t_1}, t_1)}{\|\mathbf{v}_{\theta}(\hat{\mathbf{x}}_{t_0,t_1}, t_1)\|_2} - \frac{a'_{t_1} \mathbf{x}_* + b'_{t_1} \hat{\mathbf{x}}_{t_0,t_1}}{\|a'_{t_1} \mathbf{x}_* + b'_{t_1} \hat{\mathbf{x}}_{t_0,t_1}\|_2} \right\|^2 \right].
 \end{aligned}
\end{equation}
This formulation preserves the standard flow-matching framework while explicitly guiding the predicted direction back to the true data distribution $\mathbf{x}_*$. Importantly, the inference velocity retains the correct magnitude scale inherited from the primary flow-matching loss term. As a result, integrating this velocity from $0$ to $1$ yields meaningful and highly stable generation, culminating in the best overall performance demonstrated in Tab.~\ref{tab:anti-drift_target_normalization}.

\noindent\textbf{Compatibility analysis: Anti-Drift vs. Direct State-Correction.}
We further investigate whether the model can simply learn a target direction that directly returns to the forward trajectory at timestep $t_1$. Specifically, we formulate a direct state-correction direction pointing from the drifted state to the forward perturbed state, $(\mathbf{x}_{t_1} - \hat{\mathbf{x}}_{t_0,t_1})$, and apply unit-length normalization. The results are reported in the fourth row of Tab.~\ref{tab:anti-drift_target_normalization} ("Directly Straight"). 

This experiment not only highlights the necessity of our specific \ourdri{} target but also reveals a fundamental flaw in naive correction: directly learning a path-returning correction vector severely conflicts with the primary flow-matching objective. Our proposed Anti-Drift direction remains perfectly compatible. Under an ideal scenario where perfect prediction is achieved (i.e., $\hat{\mathbf{x}}_{t_0,t_1} = \mathbf{x}_{t_1}$), the \ourdri{} target naturally degenerates back to the original flow-matching target $a'_t\mathbf{x}_*+b'_t\pmb{\epsilon}$. Conversely, under the same perfect conditions, the direct state-correction target collapses to a zero vector $\mathbf{0}$. This zero-vector target directly contradicts the flow-matching objective, causing severe gradient conflicts during training and ultimately leading to suboptimal performance.

\begin{table}[t]
\centering
\caption{\textbf{Effectiveness of exposure bias for frequency compensation.} We compare different weighting strategies for frequency compensation, including low-pass, high-pass, original-image, original-noise, and exposure bias weights. Results are reported using a SiT-B/4 model trained for 250k iterations on the CIFAR-10 conditional generation task without Classifier-Free Guidance (CFG).}
\resizebox{0.8\textwidth}{!}{
\begin{tabular}{lcccc}
\toprule
\multirow{2.5}{*}{\textbf{Method}} & \multirow{2.5}{*}{\textbf{Weight Type}} &\multirow{2.5}{*}{\textbf{Objective}} & \multicolumn{2}{c}{\textbf{CIFAR-10 (cond)}}\\  
\cmidrule(lr){4-5}
 &&& FID$\downarrow$ & IS$\uparrow$\\ 
\midrule
% SiT \cite{SiT2024} &/ & $\mathcal{L}_{FM}$ & 12.55 & 8.22 \\
\ourall{} w/o \ourfre{} + Low Pass & Low Pass & on $\mathcal{L}_{FM}$ & \underline{11.94} & \underline{8.29} \\
\ourall{} w/o \ourfre{} + High Pass & High Pass & on $\mathcal{L}_{FM}$ & 13.14 & 8.06 \\
\ourall{} w/o \ourfre{} + Original Image &$\mathbf{x}_*$ & on $\mathcal{L}_{FM}$ & 12.49 & 8.20 \\
\ourall{} w/o \ourfre{} + Original Noise &$\pmb{\epsilon}$ & on $\mathcal{L}_{FM}$ & 13.66 & 8.04 \\
\ourall{} (ours) & Exposure Bias & on $\mathcal{L}_{FM}$ & \textbf{11.22} & \textbf{8.31} \\
\bottomrule
\end{tabular}
}
\label{tab:app_exposure_bias_effectiveness}
\end{table}

\subsectionwithouttoc{Effectiveness of Exposure Bias for Frequency Compensation}
The analysis in Sec.~4.3 demonstrates that the frequency deficiency of the model varies significantly across timesteps: it lacks low-frequency components during high-noise timesteps, but accumulates excessive low-frequency content during low-noise timesteps. As shown in Tab.~\ref{tab:app_exposure_bias_effectiveness}, methods that directly inject fixed low- or high-frequency components of the original image (Rows 1 and 2), or inject the original image or noise directly (Rows 3 and 4), only partially correct the deficiency while exacerbating the opposing frequency imbalance of the model. 

In contrast, our exposure bias formulation provides a dynamic signal that accurately reflects the instantaneous frequency shortfall of the model. It adaptively supplements low-frequency information when the model lacks it during high-noise timesteps, and naturally decreases this supplementation when low-frequency components become dominant during low-noise timesteps. This adaptive behavior leads to the best overall performance, as demonstrated in Tab.~\ref{tab:app_exposure_bias_effectiveness}.

It is also worth noting that injecting exclusively low-frequency information consistently outperforms injecting exclusively high-frequency information (e.g., low-pass vs.\ high-pass, original image vs.\ original noise). This indicates that low-frequency learning has a more substantial impact on the final generative performance. Since exposure bias dynamically compensates for the specific low-frequency content the model lacks, these results further highlight the effectiveness of exposure bias as a frequency-aware corrective signal.

\begin{table}[t]
\centering
\caption{\textbf{Hyperparameter robustness across datasets and backbones.} Results are reported using 50 NFEs without Classifier-Free Guidance (CFG). Across perturbed hyperparameter settings, \ourall{} consistently improves over SiT on ImageNet-256 and CIFAR-10. We  include a SiT-M/2 setting with a deliberately suboptimal $\beta_2$.}
\resizebox{\textwidth}{!}{
\begin{tabular}{lccccc ccccc ccccc}
\toprule
\multirow{2.5}{*}{\textbf{Method}} & \multirow{2.5}{*}{\textbf{Backbone}} & \multirow{2.5}{*}{\textbf{NFE}} & \multirow{2.5}{*}{$\boldsymbol{\beta_1}$} & \multirow{2.5}{*}{$\boldsymbol{\beta_2}$} & \multirow{2.5}{*}{$\boldsymbol{\alpha}$} & \multicolumn{5}{c}{\textbf{ImageNet-256}} & \multicolumn{5}{c}{\textbf{CIFAR-10}}\\  
\cmidrule(lr){7-11}\cmidrule(lr){12-16}
 &&&&&& Hour & FID$\downarrow$ & $\Delta$FID & IS$\uparrow$ & $\Delta$IS & Hour & FID$\downarrow$ & $\Delta$FID & IS$\uparrow$ & $\Delta$IS\\ 
\midrule
SiT & SiT-B/4 & 50 & - & - & - & 166 & 61.64 & \textcolor[HTML]{A0A0A0}{-0.00} & 24.67 & \textcolor[HTML]{A0A0A0}{+0.00} & 11 & 14.09 & \textcolor[HTML]{A0A0A0}{-0.00} & 8.22 & \textcolor[HTML]{A0A0A0}{+0.00}\\
\ourall{} & SiT-B/4 & 50 & 1 & 1.0 & 1.0 & 144 & \underline{60.69} & \textcolor[HTML]{CC0000}{-0.95} & \underline{25.85} & \textcolor[HTML]{CC0000}{+1.18} & 10 & \underline{12.49} & \textcolor[HTML]{CC0000}{-1.60} & \underline{8.26} & \textcolor[HTML]{CC0000}{+0.04}\\
\ourall{} & SiT-B/4 & 50 & 10 & 1.0 & 1.0 & 144 & \textbf{60.40} & \textcolor[HTML]{CC0000}{-1.24} & \textbf{26.21} & \textcolor[HTML]{CC0000}{+1.54} & 10 & \textbf{12.01} & \textcolor[HTML]{CC0000}{-2.08} & \textbf{8.27} & \textcolor[HTML]{CC0000}{+0.05}\\
\ourall{} & SiT-B/4 & 50 & 50 & 1.0 & 1.0 & 144 & 60.83 & \textcolor[HTML]{CC0000}{-0.81} & 25.71 & \textcolor[HTML]{CC0000}{+1.04} & 10 & 12.71 & \textcolor[HTML]{CC0000}{-1.38} & \underline{8.26} & \textcolor[HTML]{CC0000}{+0.04}\\
\ourall{} & SiT-B/4 & 50 & 10 & 0.5 & 1.0 & 144 & 61.13 & \textcolor[HTML]{CC0000}{-0.51} & 25.23 & \textcolor[HTML]{CC0000}{+0.56} & 10 & 13.36 & \textcolor[HTML]{CC0000}{-0.73} & 8.24 & \textcolor[HTML]{CC0000}{+0.02}\\
\ourall{} & SiT-B/4 & 50 & 10 & 5.0 & 1.0 & 144 & 61.34 & \textcolor[HTML]{CC0000}{-0.30} & 24.91 & \textcolor[HTML]{CC0000}{+0.24} & 10 & 13.78 & \textcolor[HTML]{CC0000}{-0.31} & 8.21 & \textcolor[HTML]{009900}{-0.01}\\
\ourall{} & SiT-B/4 & 50 & 10 & 1.0 & 0.5 & 144 & 61.21 & \textcolor[HTML]{CC0000}{-0.43} & 25.09 & \textcolor[HTML]{CC0000}{+0.42} & 10 & 13.54 & \textcolor[HTML]{CC0000}{-0.55} & 8.23 & \textcolor[HTML]{CC0000}{+0.01}\\
\ourall{} & SiT-B/4 & 50 & 10 & 1.0 & 5.0 & 144 & 61.02 & \textcolor[HTML]{CC0000}{-0.62} & 25.44 & \textcolor[HTML]{CC0000}{+0.77} & 10 & 13.17 & \textcolor[HTML]{CC0000}{-0.92} & 8.25 & \textcolor[HTML]{CC0000}{+0.03}\\
\midrule
SiT & SiT-M/2 & 50 & - & - & - & 311 & 30.57 & \textcolor[HTML]{A0A0A0}{-0.00} & 58.66 & \textcolor[HTML]{A0A0A0}{+0.00} & - & - & - & - & -\\
\ourall{} & SiT-M/2 & 50 & 10 & 5.0 & 1.0 & 330 & \textbf{30.11} & \textcolor[HTML]{CC0000}{-0.46} & \textbf{58.82} & \textcolor[HTML]{CC0000}{+0.16} & - & - & - & - & -\\
\bottomrule
\end{tabular}
}
\label{tab:app_all_hyper}
\end{table}

\subsectionwithouttoc{Hyperparameter Ablation of Our Methods}
The results in Tab.~\ref{tab:app_all_hyper} show that \ourall{} remains robust under perturbed hyperparameters. On ImageNet-256, \ourall{} improves over the SiT-B/4 baseline by \textbf{0.30--1.24} FID across the tested settings, while on CIFAR-10 it improves FID by \textbf{0.31--2.08}. These results indicate that the gains are not restricted to a single carefully tuned configuration.

The same trend also holds beyond the default backbone. Even with a deliberately suboptimal setting ($\beta_2=5.0$), \ourall{} improves the SiT-M/2 baseline by \textbf{0.46} FID on ImageNet-256. Therefore, we use $\beta_1=10$, $\beta_2=1.0$, and $\alpha=1.0$ as the default configuration in our main experiments, while the ablation confirms that moderate perturbations of these coefficients preserve consistent improvements.

\subsectionwithouttoc{Multi-Step Scheduled Sampling Training for \ourdri{}}
In our main experiments, we demonstrated that simulating a single inference step during training already yields substantial improvements. We further investigate whether extending this training procedure to multi-step inference provides additional gains. Recognizing that increasing the number of simulated inference steps inherently raises the computational cost, we specifically extend our baseline to two- and three-step variants for this ablation. 

Regarding the timestep selection, we evaluate two strategies. In the \textbf{Random} timestep sampling strategy, we first randomly sample an initial timestep $t_0 \in [0,1]$, then sample the subsequent timestep $t_1$ uniformly within $(t_0, 1]$, and continue this process recursively for additional steps. In contrast, the \textbf{Equal} strategy first samples $t_0$, but then explicitly partitions the remaining sub-interval $(t_0, 1]$ into $n$ equal segments for an $n$-step simulation. Each subsequent timestep is then sampled randomly within its designated sub-interval. This mechanism enforces a more uniform temporal distribution, ensuring more stable timestep sampling compared to the fully random recursive scheme.

As shown in Rows 2 and 3 of Tab.~\ref{tab:app_multi_training}, two-step training can further improve generative performance over the one-step baseline (our main setting), provided a stable sampling strategy is used. However, extending the simulation to three steps leads to severe performance degradation (Rows 4 and 5). An explanation for this phenomenon is that excessively unrolling the simulated trajectories during training exacerbates the optimization difficulty. It compounds the variance of the gradients backpropagated through the intermediate states, making it exceedingly difficult for the network to learn consistent transitions. Comparing the sampling strategies, the Equal strategy consistently outperforms the Random strategy, confirming its superior stability.

\begin{table}[t]
\centering
\caption{\textbf{Effect of multi-step scheduled sampling for \ourdri{}.} Results are reported for unconditional generation on CIFAR-10 with 50 NFEs.}
\resizebox{0.8\textwidth}{!}{
\begin{tabular}{lccccc}
\toprule
\multirow{2.5}{*}{\textbf{Method}} & \multirow{2.5}{*}{\textbf{Steps}} & \multirow{2.5}{*}{\textbf{Sampling Strategy}} &\multirow{2.5}{*}{\textbf{Objective}} & \multicolumn{2}{c}{\textbf{CIFAR-10 (uncond)}}\\  
\cmidrule(lr){5-6}
 &&&& FID$\downarrow$ & IS$\uparrow$\\ 
\midrule
% SiT \cite{SiT2024} & / &/ & $\mathcal{L}_{FM}$ & 17.41 & 7.82 \\
\ourall{} w/o \ourfre{} & 1 & Random & Regularizer & \underline{15.92} & 7.83 \\
\ourall{} w/o \ourfre{} & 2 & Random & Regularizer & 21.89 & \textbf{8.10} \\
\ourall{} w/o \ourfre{} & 2 & Equal & Regularizer & \textbf{15.33} & \underline{8.04} \\
\ourall{} w/o \ourfre{} & 3 & Random & Regularizer & 148.95 & 3.76 \\
\ourall{} w/o \ourfre{} & 3 & Equal & Regularizer & 32.20 & 7.30 \\
\bottomrule
\end{tabular}
}
\label{tab:app_multi_training}
\end{table}

\subsection{Multi-NFE Generation on ImageNet-256 with CFG}
\begin{table}[t]
\centering
\caption{\textbf{Effect of varying inference NFEs on ImageNet-256 with CFG.} Results are reported using a SiT-B/4 model trained for 500k iterations on the conditional generation task. During training, the simulated trajectory is unrolled with 50 NFEs.}
\resizebox{0.8\textwidth}{!}{
\begin{tabular}{lcccccccccccc}
\toprule
\multirow{2.5}{*}{\textbf{Steps}} & \multicolumn{2}{c}{\textbf{SiT}} &\multicolumn{2}{c}{\textbf{IP}} & \multicolumn{2}{c}{\textbf{SDSS}} & \multicolumn{2}{c}{\textbf{MDSS}} & \multicolumn{2}{c}{\textbf{Ours(250k)}} & \multicolumn{2}{c}{\textbf{Ours(500k)}}\\  
\cmidrule(lr){2-3} \cmidrule(lr){4-5} \cmidrule(lr){6-7} \cmidrule(lr){8-9} \cmidrule(lr){10-11} \cmidrule(lr){12-13}
 & FID$\downarrow$ & IS$\uparrow$ & FID$\downarrow$ & IS$\uparrow$ & FID$\downarrow$ & IS$\uparrow$ & FID$\downarrow$ & IS$\uparrow$ & FID$\downarrow$ & IS$\uparrow$ & FID$\downarrow$ & IS$\uparrow$\\ 
\midrule
30 & 32.62 & 55.64 & \underline{30.30} & \textbf{59.14} & 31.48 & \underline{58.24} & 33.05 & 56.61 &30.36& 56.31 & \textbf{29.36} & 56.43 \\
50 & 31.64 & 55.65 & 29.84 & \textbf{60.20} & 30.29 & 58.69 & 31.90 & 56.89 &\underline{29.81}& 58.81 & \textbf{28.34} & \underline{58.94} \\
100 & 31.14 & 55.51 & 28.76 & \underline{59.04} & 29.66 & 58.66 & 31.20 & 56.93 &\underline{28.69} &59.01 & \textbf{28.05} & \textbf{59.39} \\
250 & 30.96 & 55.15 & 28.76 & 58.59 & 29.42 & 58.38 & 30.93 & 56.63 &\underline{28.56} &\underline{58.95} & \textbf{27.95} & \textbf{59.38} \\
500 & 30.94 & 54.98 & 28.55 & 58.32 & 29.41 & 58.14 & 30.90 & 56.57 &\underline{28.41} &\underline{58.81} & \textbf{27.87} & \textbf{59.36} \\
\bottomrule
\end{tabular}
}
\label{tab:app_multinfe_1.5cfg}
\end{table}

In addition to evaluating multi-step sampling under the CFG-free setting, we further assess our method with a Classifier-Free Guidance (CFG) scale of $1.5$ (following SiT~\cite{SiT2024}). As illustrated in Tab.~\ref{tab:app_multinfe_1.5cfg}, the observations remain highly consistent with the CFG-free scenario.

With CFG enabled, our method maintains robust stability across diverse inference NFEs and consistently achieves the lowest FID scores. While the strong baseline IP obtains a competitive Inception Score (IS) at smaller NFEs, its performance noticeably degrades as the sampling steps increase (e.g., dropping from $60.20$ at 50 NFEs to $58.32$ at 500 NFEs). In contrast, our IS steadily improves and stabilizes at higher NFEs, ultimately surpassing all baselines. 

\begin{table*}[t]
\centering
\caption{\textbf{Configurations on ImageNet-256, CelebA-64 and CIFAR-10.}}
\label{tab:app_configs_all}
\setlength{\extrarowheight}{2pt}
\resizebox{1\textwidth}{!}{
\begin{tabular}{@{} l *{6}{c} @{}}
\toprule
\textbf{Configs}
& \textbf{SiT-B/4} 
& \textbf{SiT-M/2} 
& \textbf{SiT-XL/2} 
& \textbf{SiT-XL/2+} 
& \textbf{SiT-B/4} 
& \textbf{SiT-B/4} \\
\midrule
Dataset & ImageNet-256 & ImageNet-256 & ImageNet-256 & ImageNet-256 & CelebA-64 & CIFAR-10 \\
Params (M) & 131 & 308 & 675 & 675 & 131 & 131 \\
Depth & 12 & 16 & 28 & 28 & 12 & 12 \\
Hidden dim & 768 & 1024 & 1152 & 1152 & 768 & 768 \\
Heads & 12 & 16 & 16 & 16 & 12 & 12 \\
Patch size & $4\times4$ & $2\times2$ & $2\times2$ & $2\times2$ & $4\times4$ & $4\times4$ \\
\midrule
Training iterations & 500k & 500k & 500k & 700k & 250k & 250k \\
Global batch size & 256 & 256 & 256 & 256 & 256 & 128 \\
Dropout & \multicolumn{6}{c}{0.0} \\
Optimizer & \multicolumn{6}{c}{Adam~\cite{adam2014}} \\
Lr schedule & \multicolumn{6}{c}{constant} \\
Learning rate & \multicolumn{6}{c}{0.0001} \\
Adam $(\beta_{\text{Adam},1},\beta_{\text{Adam},2})$ & \multicolumn{6}{c}{(0.9, 0.999)} \\
Weight decay & \multicolumn{6}{c}{0.0} \\
Gradient clip & \multicolumn{6}{c}{0.2} \\
Using NPUs & 4 & 8 & 32 & 32 & 4 & 1 \\
\midrule
$\alpha$ (exposure bias coeff.) & \multicolumn{6}{c}{1} \\
$\beta_1$ (\ourdri{} loss coeff.) & \multicolumn{6}{c}{10} \\
$\beta_2$ (\ourfre{} loss coeff.) & \multicolumn{6}{c}{1} \\
\bottomrule
\end{tabular}
}
\end{table*}

\section{More Related Work}
\label{app:more_related_work}
\textbf{Flow Rectification.} Recent advances show that directly modifying flow trajectories can effectively enhance generation quality and stability. Several works aim to accelerate sampling by redesigning the flow operators. Shortcut Models \cite{shortcut2024} introduce shortcut paths to enable one-step generation, Mean Flows \cite{mean2025} approximate the expected flow for deterministic mapping, and SplitMeanFlow \cite{smeanflow2025} enforces interval-wise consistency to stabilize few-step modeling. To reduce sampling drift, Consistency Models \cite{consistency2023} and Consistent Diffusion \cite{samplingdrift2023} learn drift-free mappings via self-consistency, while Consistency Trajectory Models \cite{CTM2024,CTMdistill2024} constrain entire probability flow ODE trajectories. In contrast, CurveFlow \cite{curveflow2025} leverages curvature to guide smoother flows, FlowEdit \cite{flowedit2025} modifies flow fields for text-based editing, and Align-Your-Flow \cite{align2025} aligns student-teacher flow maps for scalable distillation. However, these methods primarily focus on acceleration, stability, or control, rather than explicitly addressing the exposure bias caused by the mismatch between training and inference in FM.

\section{Additional Experimental Details}
\label{app:more_experiments_Settings_details}
We follow the hyperparameters of Mean Flows~\cite{mean2025} for ImageNet-256, and those of IP~\cite{IP2023} for CIFAR-10 and CelebA-64, as detailed in Tab.~\ref{tab:app_configs_all}.
For ImageNet-256, we utilize a standard VAE to compress images into $32\times32\times4$ latent representations, which serve as the model inputs. We adopt SiT~\cite{SiT2024} as our primary backbone, which is built upon the ViT~\cite{vit2020} architecture and employs adaLN-Zero~\cite{DiT2023} for class conditioning. We empirically observe that inheriting the default Classifier-Free Guidance (CFG) scales directly from the SiT baseline naturally yields the optimal generative performance for our method. For CIFAR-10, the model operates directly on the $32\times32\times3$ pixel space. We evaluate both conditional and unconditional generation on CIFAR-10. The conditional setup mirrors the ImageNet-256 configuration, while for the unconditional setting, we disable class conditioning by mapping all samples to a single default class. For CelebA-64, the model directly processes $64\times64\times3$ pixel inputs for unconditional generation, similarly employing the single-class setup. Furthermore, to evaluate the scalability of our approach on high-resolution generation, we extend our experiments to ImageNet-512 using recent advanced architectures, REPA~\cite{repa2025} and DDT~\cite{ddt2025}. To adapt them for higher resolution, we fine-tune their official pre-trained ImageNet-256 checkpoints on the ImageNet-512 dataset. For these advanced models, we adhere to their default configurations. Notably, consistent with our observations on SiT, we find that applying their respective default CFG scales also produces the best results for our method. All models are implemented in PyTorch 2.1~\cite{pytorch2019}, optimized using the Adam~\cite{adam2014} optimizer, and trained on Ascend 910B NPUs equipped with 64GB of memory.

Regarding the statistical measurements, we carefully compute cumulative metrics, such as the Predicted Frequency Ratio (PFR) and the Frequency Emphasis of Loss (FEL). FEL depends on a temporal interval rather than a single isolated timestep. Specifically, to evaluate FEL at a target timestep $t_0$, we aggregate the exposure bias across all intermediate timesteps within the interval $(t_0, 1]$. This aggregated result is subsequently averaged across all evaluated samples. In all such statistical experiments, we discretize the continuous time domain $[0,1]$ into 50 uniform timesteps.

\section{Baseline Implementation Details}
\label{app:baselin_implement_details}

\noindent\textbf{SiT~\cite{SiT2024} Details.} We directly follow the default hyperparameters and training configurations of the original SiT to train the backbone models and reproduce the baseline results.

\noindent\textbf{IP~\cite{IP2023} Details.} We adopt the default perturbation scale of 0.1. The core idea of IP is to add input perturbations to simulate inference-time errors. Following the original design, we inject an additional noise term of magnitude 0.1 into the intermediate states to mimic such inference errors and improve robustness to inference drift. This straightforward modification requires only a single-line code change to the original SiT implementation.

\noindent\textbf{SDSS~\cite{multistep2024} Details.} The method highlights its most effective strategy to align predictions with the ground truth after simulating single inference step. This alignment improves passive robustness to drift accumulated during sampling. Following the paper, we implement SDSS within the SiT backbone under Flow Matching. At each iteration, we sample $t_0<t_1$ in $[0,1]$ and infer one step from $t_0$ to $t_1$. We then evaluate the velocity on the drifted state at $t_1$ and align it with the target $a'_t\mathbf{x}_* + b'_t\pmb{\epsilon}$.

\noindent\textbf{MDSS~\cite{multistep2024} Details.} The paper reports that four training-time inference steps offer the best trade-off, and we therefore adopt the same setup. We randomly sample two timesteps $t_0, t_1 \in [0,1]$ with $t_0 < t_1$, and perform single-step inference to reach the state at $t_1$. To simulate the extended four-step inference without incurring prohibitive computational overhead from recursive model forward passes, we directly accumulate the remaining three additional steps using the current trajectory formulation as an analytical approximation. Finally, we align the final prediction with the ground truth.

\section{More Qualitative Comparison on ImageNet-256 and CelebA-64}
\begin{figure*}[htbp]
    \centering
    \includegraphics[width=1\linewidth, height=0.7\textheight]{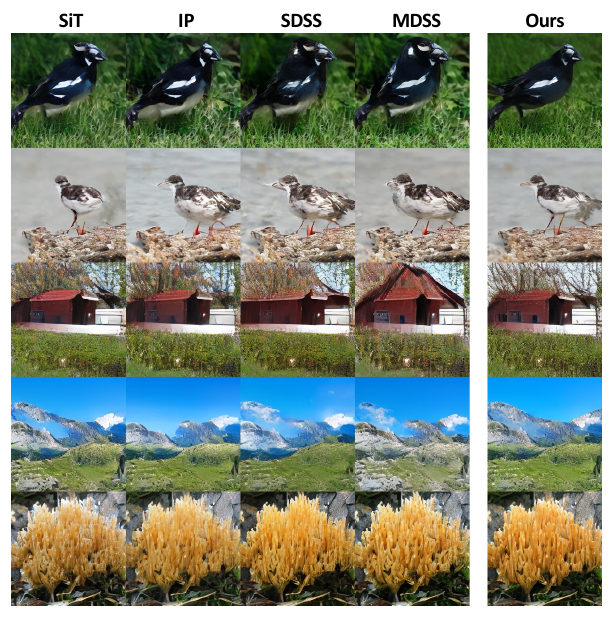}
    \caption{\textbf{Qualitative comparison across methods on ImageNet-256.} All images are generated with 50 NFEs and a CFG scale of 1.5. Compared to the baseline methods, our approach (\ourall{}) consistently yields superior visual fidelity, demonstrating more accurate semantic structures (e.g., the bird's beak in Row 1, the background trees and architecture in Row 3) and fewer visual artifacts.}
    \label{fig:app_more_qualitatvie_comparion_imagenet256}
\end{figure*}

\begin{figure*}[htbp]
    \centering
    \includegraphics[width=0.9\linewidth,height=0.8\textheight]{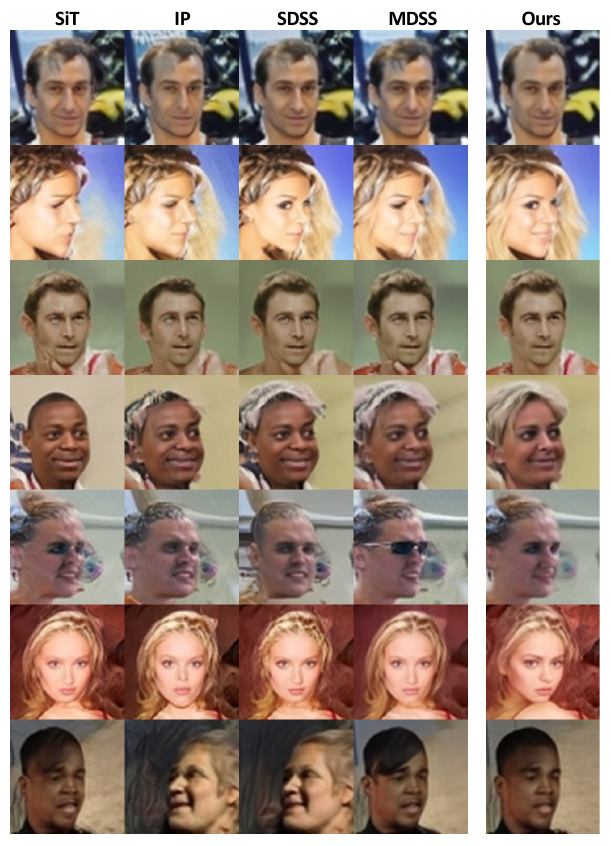}
    \caption{\textbf{Qualitative comparison across methods on CelebA-64.} All images are unconditionally generated with 50 NFEs. Compared to the baseline methods, our approach (\ourall{}) consistently yields superior visual fidelity, demonstrating cleaner facial details, more coherent hair structures, and significantly fewer artifacts (e.g., notice the forehead patches in Row 1 and the distorted facial contours in Row 5).}
    \label{fig:app_more_qualitatvie_comparion_celebA64}
\end{figure*}

In this section, we qualitatively compare \ourall{} with several strong baselines (SiT, IP, SDSS, and MDSS). To ensure a strictly fair comparison, we use the exact same random seeds and initialize the sampling trajectories with identical starting noise across all models. As shown in Fig.~\ref{fig:app_more_qualitatvie_comparion_imagenet256} and Fig.~\ref{fig:app_more_qualitatvie_comparion_celebA64}, images generated by \ourall{} consistently exhibit superior visual fidelity and structural coherence compared to those produced by the baselines.

In Fig.~\ref{fig:app_more_qualitatvie_comparion_imagenet256}, baseline models frequently fail to render key semantic structures accurately. For instance, they struggle with the bird's feet and beak, the background trees, and the fine shading on the mushroom, often leading to overexposed patches or incomplete details. In stark contrast, \ourall{} generates sharper boundaries, more coherent textures, and more faithfully reconstructed object parts.

Similarly, in Fig.~\ref{fig:app_more_qualitatvie_comparion_celebA64}, the baseline models struggle to produce clean facial regions, leaving dark patches on the forehead, blurred or inconsistent hair structures, and incorrect semantic separation between the neck and clothing. They also exhibit severe artifacts around the facial contours and fail to maintain consistent geometry between the head and neck. Across all rows, \ourall{} produces significantly cleaner facial details, more accurate hair topology, clearer background-to-foreground transitions, and drastically reduced color artifacts, culminating in a higher overall visual quality.

\section{Generated Samples on ImageNet-256 with 50 NFEs}
\label{app:more_generated_samples}
\begin{figure*}[htbp]
    \centering
    \includegraphics[width=1\linewidth,height=0.65\textheight]{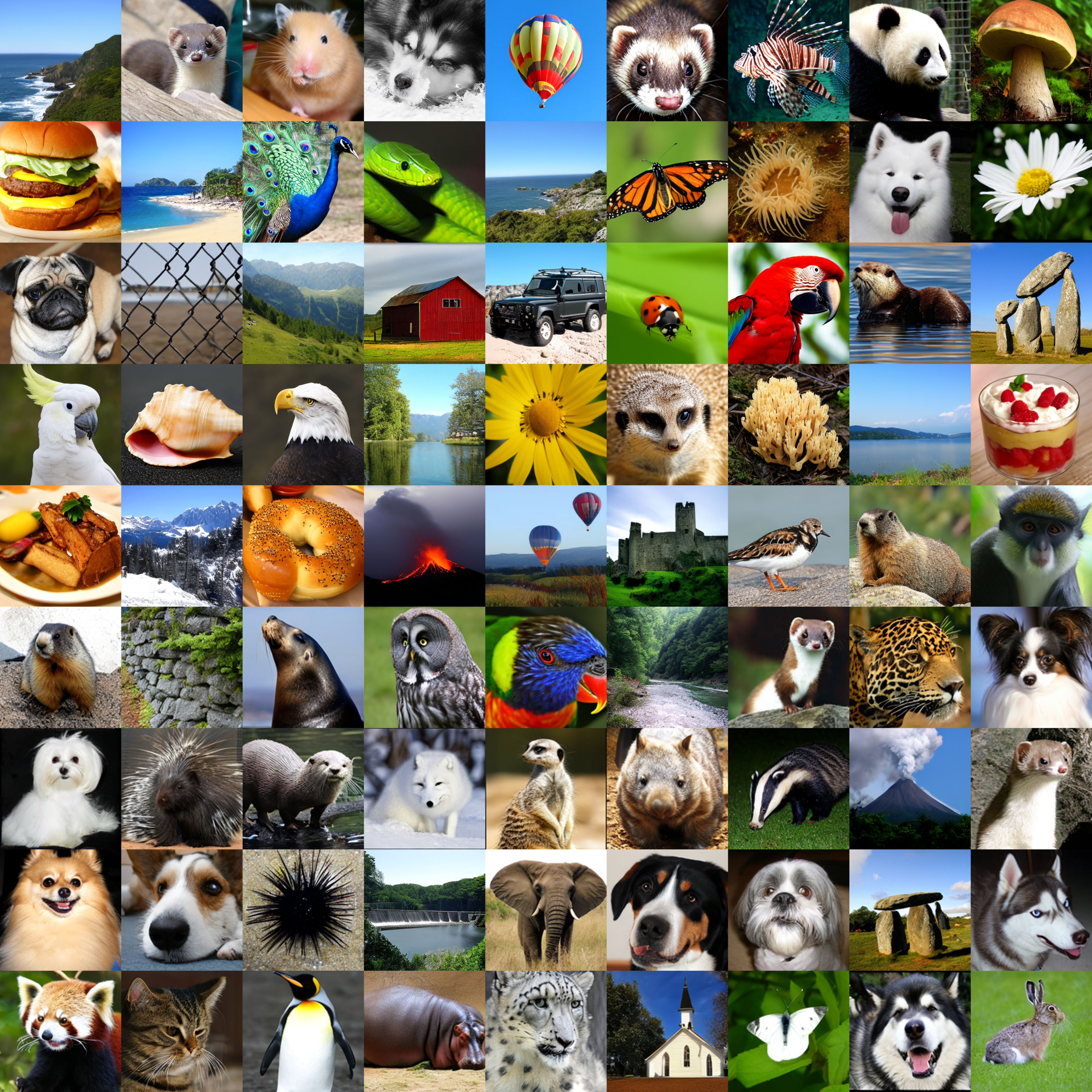}
    \caption{\textbf{Qualitative results on ImageNet-256.} The samples are generated by our \ourall{}-XL/2+ model with 50 NFEs and a CFG scale of 4.0. The results demonstrate the model's exceptional capability to synthesize structurally complex scenes and fine-grained details (e.g., intricate animal fur and natural reflections) with high visual fidelity.}
    \label{fig:app_more_generated_samples}
\end{figure*}

In this section, we present the images sampled by \ourall{} to qualitatively evaluate the generative capability of our model, as shown in Fig.~\ref{fig:app_more_generated_samples}. We deploy our best-performing \ourall{}-XL/2+ model and perform sampling with 50 NFEs and a CFG scale of 4.0 on the ImageNet-256 dataset. Across both landscapes and animal categories, the model produces exceptionally high-quality samples with fine-grained details. In many cases, \ourall{} synthesizes realistic animal characteristics, including intricate fur textures, accurate eye reflections, and subtle motion cues. It also captures structurally complex and challenging landscapes, such as volcanic eruptions, with rich structural and textural fidelity.

\section{More Frequency-Aware Visualization on ImageNet-256}
\label{app:more_awared_frequency}
\begin{figure*}[htbp]
    \centering
    \includegraphics[width=1\linewidth]{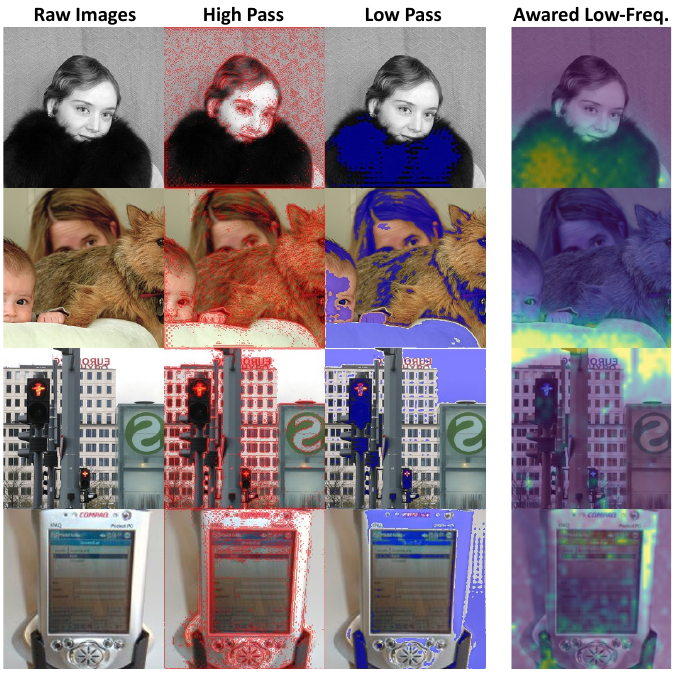}
    \caption{\textbf{Exposure bias highlights low-frequency structures in images.}
We analyze the frequency components of raw images as references and observe that exposure bias consistently aligns with low-pass regions. These examples are from ImageNet-256.}
    \label{fig:app_more_awared_frequency}
\end{figure*}

In this section, we provide detailed visual examples illustrating that exposure bias inherently acts as a frequency-aware signal, specifically attending to low-frequency regions in the original data. We employ a SiT-B/4 model trained for 500k iterations and compute the exposure bias under high-noise timesteps for various samples. We then directly visualize the resulting exposure bias spatial maps, as shown in Fig.~\ref{fig:app_more_awared_frequency}.

The leftmost column of Fig.~\ref{fig:app_more_awared_frequency} displays the original ground-truth images. Applying standard high-pass and low-pass filters to these images yields their respective high-frequency and low-frequency components (second and third columns). The rightmost column presents the visualization of the raw exposure bias maps. Across the samples, the exposure bias consistently highlights core low-frequency regions, including dark clothing, sofas and nearby objects, sky regions in outdoor scenes, and the smooth surface of the phone casing. 

These results provide visual validation that exposure bias inherently attends to semantically meaningful low-frequency pixels during high-noise timesteps, dynamically guiding the network to compensate for low-frequency deficiencies.


%%%%%%%%%%%%%%%%%%%%%%%%
\clearpage

\end{document}